# Group-based Sparse Representation for Image Restoration

Jian Zhang, *Student Member*, *IEEE*, Debin Zhao, *Member*, *IEEE*, and Wen Gao, *Fellow*, *IEEE*

**Abstract**

Traditional patch-based sparse representation modeling of natural images usually suffer from two problems. First, it has to solve a large-scale optimization problem with high computational complexity in dictionary learning. Second, each patch is considered independently in dictionary learning and sparse coding, which ignores the relationship among patches, resulting in inaccurate sparse coding coefficients. In this paper, instead of using patch as the basic unit of sparse representation, we exploit the concept of group as the basic unit of sparse representation, which is composed of nonlocal patches with similar structures, and establish a novel sparse representation modeling of natural images, called group-based sparse representation (GSR). The proposed GSR is able to sparsely represent natural images in the domain of group, which enforces the intrinsic local sparsity and nonlocal self-similarity of images simultaneously in a unified framework. Moreover, an effective self-adaptive dictionary learning method for each group with low complexity is designed, rather than dictionary learning from natural images. To make GSR tractable and robust, a split Bregman based technique is developed to solve the proposed GSR-driven $\ell_0$ minimization problem for image restoration efficiently. Extensive experiments on image inpainting, image deblurring and image compressive sensing recovery manifest that the proposed GSR modeling outperforms many current state-of-the-art schemes in both PSNR and visual perception.

*Index Terms*—Image restoration, sparse representation, nonlocal self-similarity, deblurring, inpainting, compressive sensing

J. Zhang and D. Zhao are with the School of Computer Science and Technology, Harbin Institute of Technology, Harbin 150001, China (e-mail: jzhangcs@hit.edu.cn; dbzhao@hit.edu.cn).

W. Gao is with the National Engineering Laboratory for Video Technology, and Key Laboratory of Machine Perception (MoE), School of Electrical Engineering and Computer Science, Peking University, Beijing 100871, China (e-mail: wgao@pku.edu.cn).

## I. INTRODUCTION

Image restoration has been extensively studied in the past two decades [1]–[20], whose purpose is to reconstruct the original high quality image $x$ from its degraded observed version $y$. It is a typical ill-posed linear inverse problem and can be generally formulated as:

$$y = Hx + n, \tag{1}$$

where $x, y$ are lexicographically stacked representations of the original image and the degraded image, respectively, $H$ is a matrix representing a non-invertible linear degradation operator and $n$ is usually additive Gaussian white noise. When $H$ is a mask, that is, $H$ is a diagonal matrix whose diagonal entries are either 1 or 0, keeping or killing the corresponding pixels, the problem (1) becomes image inpainting [5], [6]; when $H$ is a blur operator, the problem (1) becomes image deblurring [9], [18]; when $H$ is a set of random projections, the problem (1) becomes compressive sensing (CS) [19], [42].

To cope with the ill-posed nature of image restoration, image prior knowledge is usually employed for regularizing the solution to the following minimization problem [8]–[18]:

$$\operatorname{argmin}_x \tfrac{1}{2}\|Hx - y\|_2^2 + \lambda \Psi(x), \tag{2}$$

where $\tfrac{1}{2}\|Hx - y\|_2^2$ is the $\ell_2$ data-fidelity term, $\Psi(x)$ is called the regularization term denoting image prior and $\lambda$ is the regularization parameter. Many optimization approaches for the above regularization-based image inverse problems have been developed [16]–[18], [43].

Due to that image prior knowledge plays a critical role in the performance of image restoration algorithms, designing effective regularization terms to reflect the image priors is at the core of image restoration. Classical regularization terms, such as half quadrature formulation [21], Mumford-Shah (MS) model [22], and total variation (TV) models [1] [4], utilize local structural patterns and are built on the assumption that images are locally smooth except at edges. These regularization terms demonstrate high effectiveness in preserving edges and recovering smooth regions. However, they usually smear out image details and cannot deal well with fine structures.

In the past several years, sparsity has been emerging as one of the most significant properties of natural images [23], [24] and the sparsity-based regularization has achieved great success in various image processing applications, such as denoising [25], deblurring [11], and super-resolution [26]. The sparse model assumes that each patch of an image can be accurately represented by a few elements from a basis set called a dictionary, which is learned from natural images. Compared with traditional analytically-designed dictionaries, such as wavelets, curvelets, and bandlets, the learned dictionary enjoys the advantage of being better adapted to the images, thereby enhancing the sparsity and showing impressive performance improvement. However, there exist two main problems in the current patch-based sparse representation model. First, dictionary learning is a large-scale and highly non-convex problem, which often requires high computational complexity [24], [27]. Second, patch is the unit of sparse representation, and each patch is usually considered independently in dictionary learning and sparse coding, which ignores the relationships between similar patches in essence, such as self-similarity. Moreover, with the learned dictionary, the actual sparse coding process is always calculated with relatively expensive nonlinear estimations, such as match pursuits [28], [38], which also may be unstable and imprecise due to the coherence of the dictionary [37].

Another alternative significant property exhibited in natural images is the well-known nonlocal self-similarity, which depicts the repetitiveness of higher level patterns (e.g., textures and structures) globally positioned in images. Inspired by the success of nonlocal means (NLM) denoising filter [8], a series of nonlocal regularization terms for inverse problems exploiting nonlocal self-similarity property of natural images are emerging [32]–[36]. Due to the utilization of self-similarity prior by adaptive nonlocal graph, nonlocal regularization terms produce superior results over the local ones, with sharper image edges and more image details [33], [36]. Nonetheless, there are still plenty of image details and structures that cannot be recovered accurately. One of the reasons is that the weighted graphs adopted by the above nonlocal regularization terms inevitably give rise to disturbance and inaccuracy, due to the inaccurate weights [35].

In recent works, the sparsity and the self-similarity of natural images are usually combined to achieve better performance. In [11], sparsity and self-similarity are separately characterized by two regularization terms, which are incorporated together into the final cost functional of image restoration solution to enhance the image quality. In [12], simultaneous sparse coding is utilized to impose that similar patches should share the same dictionary elements in their sparse decomposition, which acquired impressive denoising and demosaicking results. In [15], a nonlocally centralized sparse representation (NCSR) model is proposed, which first obtains good estimates of the sparse coding coefficients of the original image by the principle of NLM in the domain of sparse coding coefficients, and then centralizes the sparse coding coefficients of the observed image to those estimates to improve the performance of sparse representation based image restoration.

Lately, low-rank modeling based approaches have also achieved great success in image or video restoration. To remove the defects in a video, unreliable pixels in the video are first detected and labeled as missing. Similar patches are grouped such that the patches in each group share similar underlying structure and form a low-rank matrix approximately. Finally, the matrix completion is carried out on each patch group to restore the image [50] [51]. In [5], a low-rank approach toward modeling nonlocal similarity denoted by SAIST is proposed, which not only provides a conceptually simple interpretation for simultaneous sparse coding [12] from a bilateral variance estimation perspective, but also achieves highly competent performance to several state-of-the-art methods.

In this paper, instead of using patch as the basic unit of sparse representation, we exploit the concept of group as the basic unit of sparse representation, and establish a novel sparse representation modeling of natural images, called group-based sparse representation (GSR). Compared with traditional patch-based sparse representation, the contributions of our proposed GSR modeling are mainly three folds. First, GSR explicitly and effectively characterizes the intrinsic local sparsity and nonlocal self-similarity of natural images simultaneously in a unified framework, which adaptively sparsifies the natural image in the domain of group. Second, an effective self-adaptive group dictionary learning method with low complexity is designed, rather than dictionary learning from natural images. Third, an efficient split Bregman based iterative algorithm is developed to solve the proposed GSR-driven $\ell_0$ minimization problem for image restoration. Experimental results on three applications: image inpainting, deblurring and image CS recovery have shown that the proposed GSR model outperforms many current state-of-the-art schemes. Part of our previous work for image CS recovery via GSR has been presented in [47].

The remainder of this paper is organized as follows. Traditional patch-based sparse representation is introduced in Section II. Section III elaborates the design of group-based sparse representation (GSR) modeling, and discusses the close relationships among the GSR model, the group sparsity model and the low rank model. Section IV proposes a new objective functional formed by our proposed GSR, and gives the implementation details of solving optimization problem. Extensive experimental results are reported in Section V. In Section VI, we summarize this paper.

## II. Traditional Patch-based Sparse Representation

In literature, the basic unit of sparse representation for natural images is patch [24]–[26]. Mathematically, denote by $\boldsymbol{x} \in \mathbb{R}^N$ and $\boldsymbol{x}_k \in \mathbb{R}^{\mathcal{B}_s}$ the vector representations of the original image and an image patch of size $\sqrt{\mathcal{B}_s} \times \sqrt{\mathcal{B}_s}$ at location $k$, $k=1,2,...,n$, where $N$ and $\mathcal{B}_s$ are the size of the whole image vector and each image patch vector, respectively, and $n$ is the number of image patches. Then we have

$$\boldsymbol{x}_k = \boldsymbol{R}_k(\boldsymbol{x}), \tag{3}$$

where $\boldsymbol{R}_k(\cdot)$ is an operator that extracts the patch $\boldsymbol{x}_k$ from the image $\boldsymbol{x}$, and its transpose, denoted by $\boldsymbol{R}_k^T(\cdot)$, is able to put back a patch into the $k$-th position in the reconstructed image, padded with zeros elsewhere. Considering that patches are usually overlapped, the recovery of $\boldsymbol{x}$ from $\{\boldsymbol{x}_k\}$ becomes

$$\boldsymbol{x} = \sum\nolimits_{k=1}^{n} \boldsymbol{R}_k^T(\boldsymbol{x}_k) \,./\, \sum\nolimits_{k=1}^{n} \boldsymbol{R}_k^T(\boldsymbol{1}_{\mathcal{B}_s}), \tag{4}$$

where the notation $./$ stands for the element-wise division of two vectors, and $\mathbf{1}_{B_s}$ is a vector of size $B_s$ with all its elements being 1. Eq. (4) indicates nothing but an abstraction strategy of averaging all the overlapped patches.

For a given dictionary $\boldsymbol{D} \in \mathbb{R}^{B_s \times M}$ ( $M$ is the number of atoms in $\boldsymbol{D}$ ), the sparse coding process of each patch $\boldsymbol{x}_k$ over $\boldsymbol{D}$ is to find a sparse vector $\boldsymbol{\alpha}_k \in \mathbb{R}^M$ (i.e., most of the coefficients in $\boldsymbol{\alpha}_k$ are zero or close to zero) such that $\boldsymbol{x}_k \approx \boldsymbol{D}\boldsymbol{\alpha}_k$. Then the entire image can be sparsely represented by the set of sparse codes $\{\boldsymbol{\alpha}_k\}$. In practice, the sparse coding problem of $\boldsymbol{x}_k$ over $\boldsymbol{D}$ is usually cast as

$$\boldsymbol{\alpha}_k = \mathrm{argmin}_{\boldsymbol{\alpha}} \tfrac{1}{2} \|\boldsymbol{x}_k - \boldsymbol{D}\boldsymbol{\alpha}\|_2^2 + \lambda \|\boldsymbol{\alpha}\|_p, \tag{5}$$

where $\lambda$ is a constant, and $p$ is 0 or 1. If $p = 0$, that means the sparsity is strictly measured by the $\ell_0$-norm of $\boldsymbol{\alpha}_k$, which counts the nonzero elements in $\boldsymbol{\alpha}_k$. Nonetheless, since the problem of $\ell_0$-norm optimization is non-convex and usually NP-hard, it is often sub-optionally solved by greedy algorithms, e.g., orthogonal matching pursuit (OMP) [28]. Alternatively, if $p = 1$, the $\ell_0$-norm minimization is approximated by the convex $\ell_1$-norm, which can be efficiently solved by some recent large-scale tools [16]–[18], [38], [43].

Similar to Eq. (4), reconstructing $\boldsymbol{x}$ from its sparse codes $\{\boldsymbol{\alpha}_k\}$ is formulated:

$$\boldsymbol{x} = \boldsymbol{D} \circ \boldsymbol{\alpha} \stackrel{def}{=} \sum_{k=1}^n \boldsymbol{R}_k^T (\boldsymbol{D}\boldsymbol{\alpha}_k) . / \sum_{k=1}^n \boldsymbol{R}_k^T (\mathbf{1}_{B_s}), \tag{6}$$

where $\boldsymbol{\alpha}$ denotes the concatenation of all $\boldsymbol{\alpha}_k$, that is, $\boldsymbol{\alpha} = [\boldsymbol{\alpha}_1^T, \boldsymbol{\alpha}_2^T, ..., \boldsymbol{\alpha}_n^T]^T$. The purpose of introducing the notation $\circ$ is to exploit $\boldsymbol{D} \circ \boldsymbol{\alpha}$ to make the expression of $\sum_{k=1}^n \boldsymbol{R}_k^T (\boldsymbol{D}\boldsymbol{\alpha}_k) . / \sum_{k=1}^n \boldsymbol{R}_k^T (\mathbf{1}_{B_s})$ more convenient.

Now, considering the degraded version in Eq. (1), the regularization-based image restoration scheme utilizing traditional patch-based sparse representation model is formulated as

$$\hat{\boldsymbol{\alpha}} = \mathrm{argmin}_{\boldsymbol{\alpha}} \tfrac{1}{2} \|\boldsymbol{H}\boldsymbol{D} \circ \boldsymbol{\alpha} - \boldsymbol{y}\|_2^2 + \lambda \|\boldsymbol{\alpha}\|_p, \tag{7}$$

where $\lambda$ is the regularization parameter, and $p$ is 0 or 1. With $\hat{\boldsymbol{\alpha}}$, the reconstructed image can be expressed by $\hat{\boldsymbol{x}} = \boldsymbol{D} \circ \hat{\boldsymbol{\alpha}}$.

The heart of the sparse representation modeling lies in the choice of dictionary $\boldsymbol{D}$. In other words, how to seek the best domain to sparsify a given image? Much effort has been devoted to learning a redundant dictionary from a set of training example image patches. To be concrete, given a set of training image patches $\boldsymbol{X} = [\boldsymbol{x}_1, \boldsymbol{x}_2, ..., \boldsymbol{x}_J]$, where $J$ is the number of training image patches, the goal of dictionary learning is to jointly optimize the dictionary $\boldsymbol{D}$ and the representation coefficients matrix $\boldsymbol{\Lambda} = [\boldsymbol{\alpha}_1, \boldsymbol{\alpha}_2, ..., \boldsymbol{\alpha}_J]$ such that $\boldsymbol{x}_k \approx \boldsymbol{D}\boldsymbol{\alpha}_k$ and $\|\boldsymbol{\alpha}_k\|_p \leq L$, where $p$ is 0 or 1. This can be formulated by the following minimization problem:

$$(\hat{\boldsymbol{D}}, \hat{\boldsymbol{\Lambda}}) = \underset{\boldsymbol{D}, \boldsymbol{\Lambda}}{\operatorname{argmin}} \sum_{k=1}^{J} \|\boldsymbol{x}_k - \boldsymbol{D}\boldsymbol{\alpha}_k\|_2^2 \text{ s.t. } \|\boldsymbol{\alpha}_k\|_p \leq L, \forall k. \qquad (8)$$

Apparently, the above minimization problem in Eq. (8) is large-scale and highly non-convex even when $p$ is 1. To make it tractable and solvable, some approximation approaches, including MOD [27] and K-SVD [24], have been proposed to optimize $D$ and $\Lambda$ alternatively, leading to many state-of-the-art results in image processing. However, these approximation approaches for dictionary learning inevitably require high computational complexity.

Apart from high computational complexity, from Eq. (5) and Eq. (8), it can be noticed that each patch is actually considered independently in dictionary learning and sparse coding, which ignores the relationships between similar patches in essence, such as self-similarity [4], [11].

### III. GROUP-BASED SPARSE REPRESENTATION (GSR)

In this paper, to rectify the above problems of traditional patch-based sparse representation, we propose a novel sparse representation modeling in the unit of group instead of patch, aiming to exploit the local sparsity and the nonlocal self-similarity of natural images simultaneously in a unified framework. Each group is represented by the form of matrix, which is composed of nonlocal patches with similar structures. Thus, the proposed sparse representation modeling is named as group-based sparse representation (GSR). Moreover, an effective self-adaptive dictionary learning method for each group with low complexity is designed rather than dictionary learning from natural images, enabling the proposed GSR more efficient and effective. This section will give detailed description of GSR modeling, and elaborate the self-adaptive dictionary learning technique.

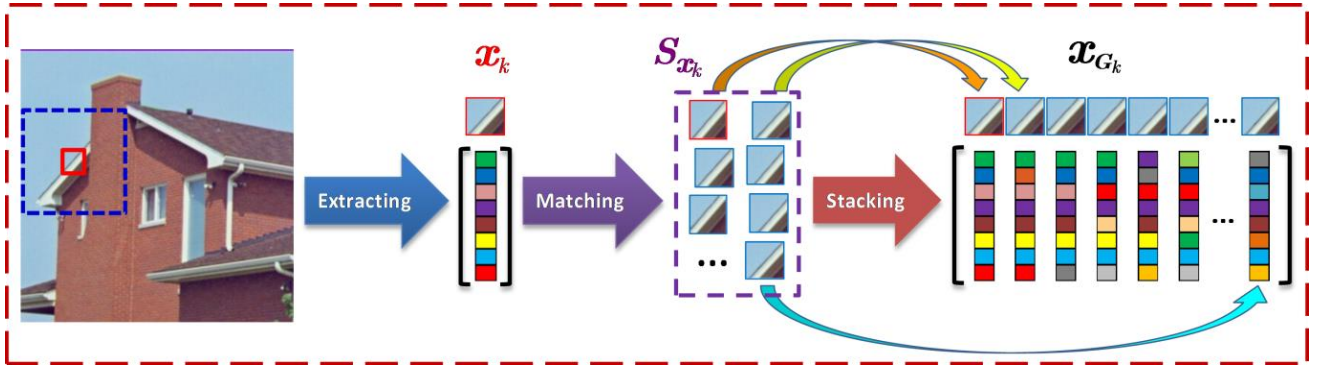

**Figure 1:** Illustrations for the group construction. Extract each patch vector $\boldsymbol{x}_k$ from image $\boldsymbol{x}$. For each $\boldsymbol{x}_k$, denote $S_{\boldsymbol{x}_k}$ the set composed of its $c$ best matched patches. Stack all the patches in $S_{\boldsymbol{x}_k}$ in the form of matrix to construct the group, denoted by $\boldsymbol{x}_{G_k}$.

#### A. Group Construction

Since the unit of our proposed sparse representation model is group, this subsection will give details to show how to construct it.

As shown in Fig. 1, first, divide the image $x$ with size $N$ into $n$ overlapped patches of size $\sqrt{B_s} \times \sqrt{B_s}$ and each patch is denoted by the vector $x_k \in \mathbb{R}^{B_s}$, i.e., $k = 1, 2, ..., n$.

Then, for each patch $x_k$, denoted by small red square in Fig. 1, in the $L \times L$ training window (big blue square), search its $c$ best matched patches, which comprise the set $S_{x_k}$. Here, Euclidean distance is selected as the similarity criterion between different patches.

Next, all the patches in $S_{x_k}$ are stacked into a matrix of size $B_s \times c$, denoted by $x_{G_k} \in \mathbb{R}^{B_s \times c}$, which includes every patch in $S_{x_k}$ as its columns, i.e., $x_{G_k} = \{x_{G_k \otimes 1}, x_{G_k \otimes 2}, ..., x_{G_k \otimes c}\}$. The matrix $x_{G_k}$ containing all the patches with similar structures is named as a group. Analogous to Eq. (3), we define

$$x_{G_k} = R_{G_k}(x), \tag{9}$$

where $R_{G_k}(\cdot)$ is actually an operator that extracts the group $x_{G_k}$ from $x$, and its transpose, denoted by $R_{G_k}^T(\cdot)$, can put back a group into the $k$-th position in the reconstructed image, padded with zeros elsewhere.

By averaging all the groups, the recovery of the whole image $x$ from $\{x_{G_k}\}$ becomes

$$x = \sum_{k=1}^{n} R_{G_k}^T(x_{G_k}) \,./\, \sum_{k=1}^{n} R_{G_k}^T(\mathbf{1}_{B_s \times c}), \tag{10}$$

where $./$ stands for the element-wise division of two vectors and $\mathbf{1}_{B_s \times c}$ is a matrix of size $B_s \times c$ with all the elements being 1.

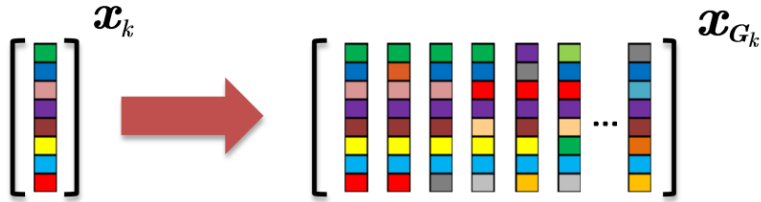

**Figure 2:** Comparison between patch $x_k$ and group $x_{G_k}$.

Note that, in our paper, each patch $x_k$ is represented as a vector, while each group $x_{G_k}$ is represented as a matrix, as illustrated in Fig. 2. According to above definition, it is obvious to observe that each patch $x_k$ corresponds to a group $x_{G_k}$. One can also see that the construction of $x_{G_k}$ explicitly exploits the self-similarity of natural images.

*B. Group-based Sparse Representation Modeling*

To enforce the local sparsity and the nonlocal self-similarity of natural images simultaneously in a unified framework, we propose to sparsify natural images in the domain of group. Therefore, our proposed model is named as group-based sparse representation (GSR). The proposed GSR model assumes that each group $x_{G_k}$ can be accurately represented by a few atoms of a self-adaptive learning dictionary $D_{G_k}$.

In this subsection, $D_{G_k} = [d_{G_k \otimes 1}, d_{G_k \otimes 2}, ..., d_{G_k \otimes m}]$ is supposed to be known. Note that each atom $d_{G_k \otimes i} \in \mathbb{R}^{B_s \times c}$ is a matrix of the same size as the group $x_{G_k}$, and $m$ is the number of atoms in $D_{G_k}$. Different from the dictionary in patch-based sparse representation, here $D_{G_k}$ is of size $(B_s \times c) \times m$, that is, $D_{G_k} \in \mathbb{R}^{(B_s \times c) \times m}$. How to learn $D_{G_k}$ with low complexity will be given in the next subsection.

Thus, some notations about GSR can be readily extended from patch-based sparse representation. Specifically, the sparse coding process of each group $x_{G_k}$ over $D_{G_k}$ is to seek a sparse vector $\alpha_{G_k} = [\alpha_{G_k \otimes 1}, \alpha_{G_k \otimes 2}, ..., \alpha_{G_k \otimes m}]$ such that $x_{G_k} \approx \sum_{i=1}^{m} \alpha_{G_k \otimes i} d_{G_k \otimes i}$. For simplicity, we utilize the expression $D_{G_k} \alpha_{G_k}$ to represent $\sum_{i=1}^{m} \alpha_{G_k \otimes i} d_{G_k \otimes i}$ without confusion. Note that $D_{G_k} \alpha_{G_k}$ is not a strict matrix-vector multiplication. It is also worth emphasizing that the sparse coding process of each group under our proposed $D_{G_k}$ is quite efficient without iteration, which will be seen in the following section. Then the entire image can be sparsely represented by the set of sparse codes $\{\alpha_{G_k}\}$ in the group domain. Reconstructing $x$ from the sparse codes $\{\alpha_{G_k}\}$ is expressed as:

$$x = D_G \circ \alpha_G \overset{def}{=} \sum_{k=1}^{n} R_{G_k}^T (D_{G_k} \alpha_{G_k}) . \Big/ \sum_{k=1}^{n} R_{G_k}^T (\mathbf{1}_{B_s \times c}), \tag{11}$$

where $D_G$ denotes the concatenation of all $D_{G_k}$, and $\alpha_G$ denotes the concatenation of all $\alpha_{G_k}$.

Accordingly, by considering the degraded version in Eq. (1), the proposed regularization-based image restoration scheme via GSR is formulated as

$$\hat{\alpha}_G = \operatorname{argmin}_{\alpha_G} \tfrac{1}{2} \| H D_G \circ \alpha_G - y \|_2^2 + \lambda \| \alpha_G \|_0. \tag{12}$$

With $\hat{\alpha}_G$, the reconstructed image can be expressed by $\hat{x} = D_G \circ \hat{\alpha}_G$. Note that, in this paper, $\ell_0$-norm is exploited to measure the real sparsity of $\alpha_G$ in the group domain in order to enhance the image restoration quality. Nonetheless, Eq. (12) is usually hard to solve owing that $\ell_0$-norm optimization is non-convex. Thus, solving Eq. (12) efficiently and effectively is one of our main contributions, which will be seen in the following.

To understand GSR model more clearly, here, we make a comparison between Eq. (12) and previous patch-based sparse representation for image restoration in Eq. (7). We can see the differences between Eq. (12) and Eq. (7) lie in the dictionary and the unit of sparse representation. The advantages of Eq. (12) are mainly three-folds. First, GSR adopts group as the unit of sparse representation and sparsely represents the entire image in the group domain. Since the group is composed of patches with similar structures, GSR exploits self-similarity explicitly in dictionary learning and sparse coding, which is more robust and effectual. Second, rather than learning a general dictionary $D$ for all patches in Eq. (7), a self-adaptive dictionary $D_{G_k}$ is designed for each

$x_{G_k}$, which is more effective. Third, as will be seen below, the proposed self-adaptive dictionary learning of $D_{G_k}$ is with low complexity, which doesn't require high computational complexity to solve large-scale optimizations.

*C. Self-Adaptive Group Dictionary Learning*

In this subsection, we will show how to learn the adaptive dictionary $D_{G_k}$ for each group $x_{G_k}$. Note that, on one hand, we hope that each $x_{G_k}$ can be faithfully represented by $D_{G_k}$. On the other hand, it is expected that the representation coefficient vector of $x_{G_k}$ over $D_{G_k}$ is as sparse as possible. Like traditional patch-based dictionary learning algorithm in Eq. (8), the adaptive dictionary learning of group can be intuitively formulated as:

$$\underset{D_x, \{\alpha_{G_k}\}}{\operatorname{argmin}} \sum_{k=1}^{n} \|x_{G_k} - D_x \alpha_{G_k}\|_F^2 + \lambda \sum_{k=1}^{n} \|\alpha_{G_k}\|_p, \quad (13)$$

where $p$ is 0 or 1. Eq. (13) is a joint optimization problem of $D_x$ and $\{\alpha_{G_k}\}$, which can be solved by alternatively optimizing $D_x$ and $\{\alpha_{G_k}\}$.

Nevertheless, we do not directly utilize Eq. (13) to learn the dictionary for each group $x_{G_k}$ based on the following three considerations. First, solving the joint optimization in Eq. (13) requires much computational cost, especially in the unit of group. Second, the learnt dictionary from Eq. (13) is actually adaptive for a given image $x$, not adaptive for a group $x_{G_k}$, which means that all the groups $\{x_{G_k}\}$ are represented by the same dictionary $D_x$. That's why the dictionary learnt by Eq. (13) here is denoted by $D_x$, rather than $\{D_{G_k}\}$. Finally, the dictionary learning process in Eq. (13) neglects the characteristics of each group $x_{G_k}$, which contains patches with similar patterns. That is to say it is not necessary to learn an over-complete dictionary, and it is even possible to learn a dictionary by a more efficient and effective manner.

Similar to the idea of dictionary learning strategy using similar patches in [3], we propose to learn the adaptive dictionary $D_{G_k}$ for each group $x_{G_k}$ directly from its estimate $r_{G_k}$, because in practice the original image $x$ is not available for learning all the groups' dictionaries $\{D_{G_k}\}$. The estimate $r_{G_k}$ will be naturally selected in the process of optimization, which will be explained in the following sections.

After obtaining $r_{G_k}$, we then apply SVD to it, that is,

$$r_{G_k} = U_{G_k} \Sigma_{G_k} V_{G_k}^T = \sum_{i=1}^{m} \gamma_{r_{G_k} \otimes i} (u_{G_k \otimes i} v_{G_k \otimes i}^T), \quad (14)$$

where $\gamma_{r_{G_k}} = [\gamma_{r_{G_k} \otimes 1}; \gamma_{r_{G_k} \otimes 2}; ...; \gamma_{r_{G_k} \otimes m}]$, $\Sigma_{G_k} = \operatorname{diag}(\gamma_{r_{G_k}})$ is a diagonal matrix with the elements of $\gamma_{r_{G_k}}$ on its main diagonal, and $u_{G_k \otimes i}, v_{G_k \otimes i}$ are the columns of $U_{G_k}$ and $V_{G_k}$, separately. Each atom in $D_{G_k}$ for group $x_{G_k}$, is defined as

$$d_{G_k \otimes i} = u_{G_k \otimes i} v_{G_k \otimes i}^T, \quad i = 1, 2, ..., m, \quad (15)$$

where $d_{G_k \otimes i} \in \mathbb{R}^{B_s \times c}$. Therefore, the ultimate adaptively learned dictionary for $x_{G_k}$ is defined as

$$D_{G_k} = [d_{G_k \otimes 1}, d_{G_k \otimes 2}, ..., d_{G_k \otimes m}]. \qquad (16)$$

According to the above definitions, the main difference between [3] and this paper for dictionary learning is that we utilize SVD to learn an adaptive dictionary for each group, while [3] utilizes PCA to learn an adaptive dictionary for each patch. The advantage of our proposed dictionary learning for each group is that it can guarantee all the patches in each group use the same dictionary and share the same dictionary atoms, which is more effective and robust, while [3] just trained the dictionary for each patch independently using its similar patches. It is clear to see that the proposed group dictionary learning is self-adaptive to each group $x_{G_k}$ and is quite efficient, requiring only one SVD for each group. In addition, owing to unitary property of $D_{G_k}$, the sparse coding process is not only efficient, but also stable and precise, which will be seen in the next section.

*D. Discussions*

This subsection will provide the detailed discussions about the close relationships among the proposed GSR model, the group sparsity model, and the low rank model. In fact, all the three models are involved with a set of similar patches to exploit the self-similarity of natural images.

As illustrated in Fig. 1, the proposed GSR model aims to adaptively seek the sparse representation of natural images in the unit of the group $x_{G_k}$. The group sparsity model imposes that similar patches in $x_{G_k}$ should share the same dictionary elements in their sparse decomposition. The low rank model hopes to find a low rank approximation of $x_{G_k}$ in order to find the robust estimate. These three models seem different at first glance, since they start from different views. However, interestingly, with the aid of the particular dictionary learning method by SVD, one type of the group sparsity model, and one type of the low rank model can be derived from our proposed GSR model, respectively. That means these three models are equivalent to some extent, which is of great help to understand these three models integrally. The details are given below.

As shown in Fig. 1, for each group $x_{G_k}$, given its noisy estimate $r_{G_k}$, the proposed GSR model to estimate $x_{G_k}$ such that $x_{G_k} = D_{G_k} \alpha_{G_k}$ is formulated as

$$\hat{\alpha}_{G_k} = \operatorname{argmin}_{\alpha_{G_k}} \tfrac{1}{2} \|r_{G_k} - D_{G_k} \alpha_{G_k}\|_F^2 + \lambda \|\alpha_{G_k}\|_0. \qquad (17)$$

With $\hat{\alpha}_{G_k}$, the reconstructed group is then expressed by $\hat{x}_{G_k} = D_{G_k} \hat{\alpha}_{G_k}$.

Assume $x_{G_k} = DA$, where $D \in \mathbb{R}^{B_s \times m}$ is the dictionary to sparsely represent all the patches in $x_{G_k}$, and $A \in \mathbb{R}^{m \times c}$ denotes the coefficient matrix. Here, set $D$ to be $U_{G_k}$ in Eq. (14), and in the light of all the definitions above, Eq. (17) is equivalent to the following form:

$$\hat{\boldsymbol{A}} = \operatorname{argmin}_{\boldsymbol{A}} \tfrac{1}{2} \|\boldsymbol{r}_{G_k} - \boldsymbol{DA}\|_F^2 + \lambda \|\boldsymbol{A}\|_{0,\infty}, \tag{18}$$

where $\|\cdot\|_{0,\infty}$ denotes the number of the nonzero rows of a matrix and is a pseudo norm [12] [29]. With $\hat{\boldsymbol{A}}$, we get $\hat{\boldsymbol{x}}_{G_k} = \boldsymbol{D}\hat{\boldsymbol{A}}$.

Due to the definition of the group sparsity model [5], [12], [29], one can see that Eq. (18) is just the special case of the group sparsity model with the constraint of the $\ell_{0,\infty}$ matrix norm, which differs from the previous group sparsity models with the constraint of the $\ell_{1,2}$ matrix norm [12] [5].

Similarly, define $\boldsymbol{\gamma}_{\boldsymbol{x}_{G_k}}$ the vector composed of all the singular values of $\boldsymbol{x}_{G_k}$, i.e., $\boldsymbol{\gamma}_{\boldsymbol{x}_{G_k}} = [\gamma_{\boldsymbol{x}_{G_k} \otimes 1}; \gamma_{\boldsymbol{x}_{G_k} \otimes 2}; \ldots; \gamma_{\boldsymbol{x}_{G_k} \otimes m}]$,. Due to $\boldsymbol{x}_{G_k} = \boldsymbol{D}_{G_k} \boldsymbol{\alpha}_{G_k}$ and the definitions of $\boldsymbol{D}_{G_k}$, we obtain

$$\|\boldsymbol{\gamma}_{\boldsymbol{x}_{G_k}}\|_0 = \operatorname{rank}(\boldsymbol{x}_{G_k}) = \|\boldsymbol{\alpha}_{G_k}\|_0, \tag{19}$$

where $\operatorname{rank}(\cdot)$ represents the rank of a matrix. Therefore, the following equation can be derived from Eq. (17):

$$\hat{\boldsymbol{x}}_{G_k} = \operatorname{argmin}_{\boldsymbol{x}_{G_k}} \tfrac{1}{2} \|\boldsymbol{x}_{G_k} - \boldsymbol{r}_{G_k}\|_F^2 + \lambda \|\boldsymbol{\gamma}_{\boldsymbol{x}_{G_k}}\|_0, \tag{20}$$

which is just the low rank model with the $\ell_0$ norm of the vector composed of all the singular values of $\boldsymbol{x}_{G_k}$ and differs from previous low rank models with the $\ell_1$ norm of the singular values vector [50] [51].

## IV. Optimization for GSR-driven $\ell_0$ minimization

In this section, an efficient approach is developed to solve the proposed GSR-driven $\ell_0$ minimization for image restoration in Eq. (12), which is one of our main contributions.

$$\hat{\boldsymbol{\alpha}}_G = \operatorname{argmin}_{\boldsymbol{\alpha}_G} \tfrac{1}{2} \|\boldsymbol{HD}_G \circ \boldsymbol{\alpha}_G - \boldsymbol{y}\|_2^2 + \lambda \|\boldsymbol{\alpha}_G\|_0. \tag{12}$$

Since $\ell_0$ minimization is non-convex and NP-hard, the usual routine is to solve its optimal convex approximation, i.e., $\ell_1$ minimization, which has been proved that, under some conditions, $\ell_1$ minimization is equivalent to $\ell_0$ minimization in a technical sense. The $\ell_1$ minimization can be solved efficiently by some recent convex optimization algorithms, such as iterative shrinkage/thresholding [16], [17], split Bregman algorithms [43]. Therefore, the straightforward method to solve Eq. (12) is translated into solving its $\ell_1$ convex form, that is

$$\hat{\boldsymbol{\alpha}}_G = \operatorname{argmin}_{\boldsymbol{\alpha}_G} \tfrac{1}{2} \|\boldsymbol{HD}_G \circ \boldsymbol{\alpha}_G - \boldsymbol{y}\|_2^2 + \lambda \|\boldsymbol{\alpha}_G\|_1. \tag{21}$$

However, a fact that is often neglected is, for some practical problems including image inverse problems, the conditions describing the equivalence of $\ell_0$ minimization and $\ell_1$ minimization are not necessarily satisfied. Therefore, this paper attempts to exploit the framework of convex optimization algorithms to solve the $\ell_0$ minimization. Experimental results demonstrate the effectiveness and the convergence of our proposed approach. The superiority of solving Eq. (12) over solving Eq. (21) is further discussed in the experimental section.

In this paper, we adopt the framework of split Bregman iteration (SBI) [43] to solve Eq. (12), which is verified to be more effective than iterative shrinkage/thresholding (IST) [16] in our experiments (See Section V for more details).

First of all, let's make a brief review of SBI. The SBI algorithm was first proposed in [43] and was shown to be powerful in for solving various variational models [43], [44], [49]. Consider a constrained optimization problem

$$\min_{u \in \mathbb{R}^N, v \in \mathbb{R}^M} f(u) + g(v), \text{ s.t. } u = Gv, \qquad (22)$$

where $G \in \mathbb{R}^{M \times N}$ and $f: \mathbb{R}^N \to \mathbb{R}$, $g: \mathbb{R}^M \to \mathbb{R}$ are convex functions. The SBI to address problem (22) works as follows:

**Algorithm 1** *Split Bregman Iteration (SBI)*

1. **Set** $t = 0$, choose $\mu > 0$, $b_0 = 0, u_0 = 0, v_0 = 0$.
2. **Repeat**
3. $u^{(t+1)} = \operatorname{argmin}_u f(u) + \frac{\mu}{2} \|u - Gv^{(t)} - b^{(t)}\|_2^2$;
4. $v^{(t+1)} = \operatorname{argmin}_v g(v) + \frac{\mu}{2} \|u^{(t+1)} - Gv - b^{(t)}\|_2^2$;
5. $b^{(t+1)} = b^{(t)} - (u^{(t+1)} - Gv^{(t+1)})$;
6. $t \leftarrow t + 1$;
7. **Until** stopping criterion is satisfied

In SBI, the parameter $\mu$ is fixed to avoid the problem of numerical instabilities instead of choosing a predefined sequence that tends to infinity as done in [30]. According to SBI, the original minimization (22) is split into two sub-problems. The rationale behind is that each sub-problem minimization may be much easier than the original problem (22).

Now, let us go back to Eq. (12) and point out how to apply the framework of SBI to solve it. By introducing a variable $u$, we first transform Eq. (12) into an equivalent constrained form,

$$\min_{\alpha_G, u} \frac{1}{2}\|Hu - y\|_2^2 + \lambda \|\alpha_G\|_0, \text{ s.t. } u = D_G \circ \alpha_G. \qquad (23)$$

Define $f(u) = \frac{1}{2}\|Hu - y\|_2^2$, $g(\alpha_G) = \lambda \|\alpha_G\|_0$.

Then, invoking SBI, Line 3 in **Algorithm 1** becomes:

$$u^{(t+1)} = \operatorname*{argmin}_{u} \frac{1}{2}\|Hu - y\|_2^2 + \frac{\mu}{2}\|u - D_G \circ \alpha_G^{(t)} - b^{(t)}\|_2^2. \qquad (24)$$

Next, Line 4 in **Algorithm 1** becomes:

$$\alpha_G^{(t+1)} = \mathrm{argmin}_{\alpha_G} \lambda \|\alpha_G\|_0 + \frac{\mu}{2}\|u^{(t+1)} - D_G \circ \alpha_G - b^{(t)}\|_2^2. \tag{25}$$

According to Line 5 in **Algorithm 1**, the update of $b^{(t)}$ is

$$b^{(t+1)} = b^{(t)} - (u^{(t+1)} - D_G \circ \alpha_G^{(t+1)}). \tag{26}$$

Thus, by SBI, the minimization for Eq. (12) is transformed into solving two sub-problems, namely, $u$, $\alpha_G$ sub-problems. In the following, we will provide the implementation details to obtain the efficient solutions to each separated sub-problem. For simplicity, the subscript $t$ is omitted without confusion.

A. $u$ Sub-problem

Given $\alpha_G$, the $u$ sub-problem denoted by Eq. (24) is essentially a minimization problem of strictly convex quadratic function, that is

$$\min_u Q_1(u) = \min_u \frac{1}{2}\|Hu - y\|_2^2 + \frac{\mu}{2}\|u - D_G \circ \alpha_G - b\|_2^2. \tag{27}$$

Setting the gradient of $Q_1(u)$ to be zero gives a closed solution for Eq. (27), which can be expressed as

$$\hat{u} = (H^T H + \mu I)^{-1} q, \tag{28}$$

where $q = H^T y + \mu(D_G \circ \alpha_G + b)$, $I$ is identity matrix.

As for image inpainting and image deblurring, due to the special structure of $H$, Eq. (28) can be computed efficiently without computing the matrix inverse (more details can be found in [18]).

As for image compressive sensing (CS) recovery, $H$ is a random projection matrix without special structure. Thus, it is too costly to solve Eq. (27) directly by Eq. (28). Here, to avoid computing the matrix inverse, the gradient descent method is utilized to solve Eq. (27) by applying

$$\hat{u} = u - \eta d, \tag{29}$$

where $d$ is the gradient direction of the objective function $Q_1(u)$ and $\eta$ represents the optimal step. Therefore, solving $u$ sub-problem for image CS recovery only requires computing the following equation iteratively

$$\hat{u} = u - \eta(H^T H u - H^T y + \mu(u - D_G \circ \alpha_G - b)), \tag{30}$$

where $H^T H$ and $H^T y$ can be calculated before, making above computation more efficient.

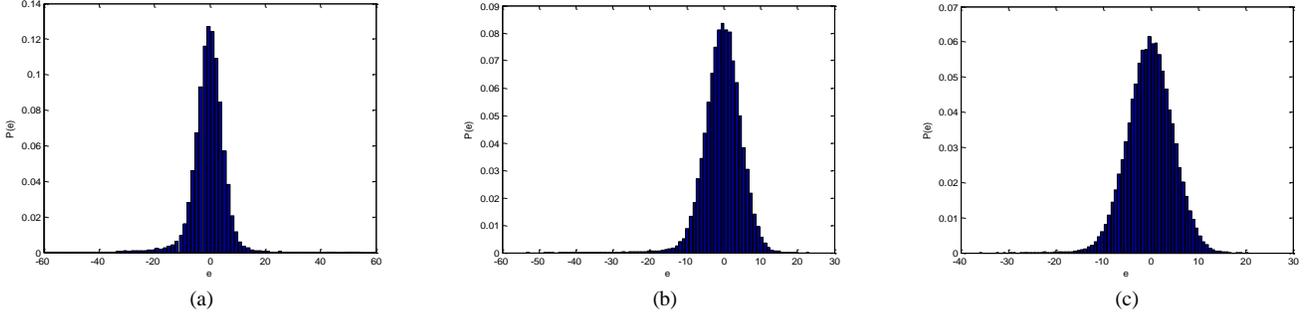

**Figure 3:** The distribution of $e^{(t)}$ and its corresponding variance $\mathrm{Var}(e^{(t)})$ for image *Parrots* in the case of image deblurring at different iterations. (a) $t = 3$ and $\mathrm{Var}(e^{(3)}) = 25.70$; (b) $t = 5$ and $\mathrm{Var}(e^{(5)}) = 23.40$; (c) $t = 7$ and $\mathrm{Var}(e^{(7)}) = 23.16$.

### B. $\alpha_G$ Sub-problem

Given $u$, according to Eq. (25), the $\alpha_G$ sub-problem can be formulated as

$$\min_{\alpha_G} Q_2(\alpha_G) = \min_{\alpha_G} \tfrac{1}{2}\|D_G \circ \alpha_G - r\|_2^2 + \tfrac{\lambda}{\mu}\|\alpha_G\|_0, \tag{31}$$

where $r = u - b$.

Note that it is difficult to solve Eq. (31) directly due to the complicated definition of $\alpha_G$. Instead, we make some transformation. Let $x = D_G \circ \alpha_G$, then Eq. (31) equally becomes

$$\min_{\alpha_G} \tfrac{1}{2}\|x - r\|_2^2 + \tfrac{\lambda}{\mu}\|\alpha_G\|_0. \tag{32}$$

By viewing $r$ as some type of noisy observation of $x$, we perform some experiments to investigate the statistics of $e = x - r$. Here, we use image *Parrots* as an example in the case of image deblurring, where the original image is first blurred by uniform blur kernel and then is added by Gaussian white noise of standard deviation 0.5. At each iteration $t$, we can obtain $r^{(t)}$ by $r^{(t)} = u^{(t)} - b^{(t-1)}$. Since the exact minimizer of Eq. (32) is not available, we then approximate $x^{(t)}$ by the original image without generality. Therefore, we are able to acquire the histogram of $e^{(t)} = x^{(t)} - r^{(t)}$ at each iteration $t$. Fig. 3 shows the distributions of $e^{(t)}$ when $t$ equals to 3, 5 and 7, respectively.

From Fig. 3, it is obvious to observe that the distribution of $e^{(t)}$ at each iteration is quite suitable to be characterized by generalized Gaussian distribution (GGD) [39] with zero-mean and variance $\mathrm{Var}(e^{(t)})$. The variance $\mathrm{Var}(e^{(t)})$ can be estimated by

$$\mathrm{Var}(e^{(t)}) = \tfrac{1}{N}\|x^{(t)} - r^{(t)}\|_2^2. \tag{33}$$

Fig. 3 also gives the corresponding estimated variances at different iterations. Furthermore, owing that the residual of images is usually de-correlated, each element of $e^{(t)}$ can be modeled independently.

Accordingly, to enable solving Eq. (32) tractable, in this paper, a reasonable assumption is made, with which even a closed-form solution of Eq. (32) can be obtained. We suppose that each element of $e^{(t)}$ follows an independent zero-mean distribution with variance $\text{Var}(e^{(t)})$. It is worth emphasizing that the above assumption does not need to be Gaussian, or Laplacian, or GGD process, which is more general. By this assumption, we can prove the following conclusion.

**THEOREM 1.** Let $x, r \in \mathbb{R}^N$, $x_{G_k}, r_{G_k} \in \mathbb{R}^{B_s \times c}$, and denote the error vector by $e = x - r$ and each element of $e$ by $e(j)$, $j = 1, ..., N$. Assume that $e(j)$ is independent and comes from a distribution with zero mean and variance $\sigma^2$. Then, for any $\varepsilon > 0$, we have the following property to describe the relationship between $\|x - r\|_2^2$ and $\sum_{k=1}^{n} \|x_{G_k} - r_{G_k}\|_F^2$, that is,

$$\lim_{\substack{N \to \infty \\ K \to \infty}} P\left\{ \left| \frac{1}{N} \|x - r\|_2^2 - \frac{1}{K} \sum_{k=1}^{n} \|x_{G_k} - r_{G_k}\|_F^2 \right| < \varepsilon \right\} = 1, \tag{34}$$

here $P(\cdot)$ represents the probability and $K = B_s \times c \times n$ (See Appendix A for detailed proof).

According to **Theorem 1**, there exists the following equation with very large probability (limited to 1) at each iteration $t$:

$$\frac{1}{N} \|x^{(t)} - r^{(t)}\|_2^2 = \frac{1}{K} \sum_{k=1}^{n} \|x_{G_k}^{(t)} - r_{G_k}^{(t)}\|_F^2. \tag{35}$$

Now let's verify Eq. (35) by the above case of image deblurring. We can clearly see that the left hand of Eq. (35) is just $\text{Var}(e^{(t)})$ defined in Eq. (33), with $\text{Var}(e^{(3)}) = 25.70$, $\text{Var}(e^{(5)}) = 23.40$, and $\text{Var}(e^{(7)}) = 23.16$, which is shown in Fig. 3. At the same time, we can calculate the corresponding right hand of Eq. (35), denoted by $\text{Var}(e_G^{(t)})$, with the same values of $t$, leading to $\text{Var}(e_G^{(3)}) = 25.21$, $\text{Var}(e_G^{(5)}) = 23.15$, and $\text{Var}(e_G^{(7)}) = 23.07$. Apparently, at each iteration, $\text{Var}(e^{(t)})$ is very close to $\text{Var}(e_G^{(t)})$, especially when $t$ is larger, which sufficiently illustrates the validity of our assumption.

Next, by incorporating Eq. (35) into Eq. (32), it yields

$$\begin{aligned}
&\min_{\alpha_G} \frac{1}{2} \sum_{k=1}^{n} \|x_{G_k} - r_{G_k}\|_F^2 + \frac{\lambda K}{\mu N} \|\alpha_G\|_0 \\
&= \min_{\alpha_G} \frac{1}{2} \sum_{k=1}^{n} \|x_{G_k} - r_{G_k}\|_F^2 + \frac{\lambda K}{\mu N} \sum_{k=1}^{n} \|\alpha_{G_k}\|_0 \\
&= \min_{\alpha_G} \sum_{k=1}^{n} \frac{1}{2} \|x_{G_k} - r_{G_k}\|_F^2 + \tau \|\alpha_{G_k}\|_0,
\end{aligned} \tag{36}$$

where $\tau = (\lambda K)/(\mu N)$.

It is obvious to see that Eq. (36) can be efficiently minimized by solving $n$ sub-problems for all the groups $x_{G_k}$. Each group based sub-problem is formulated as:

$$\operatorname{argmin}_{\boldsymbol{\alpha}_{G_k}} \frac{1}{2} \left\| \boldsymbol{x}_{G_k} - \boldsymbol{r}_{G_k} \right\|_F^2 + \tau \left\| \boldsymbol{\alpha}_{G_k} \right\|_0$$
$$= \operatorname{argmin}_{\boldsymbol{\alpha}_{G_k}} \frac{1}{2} \left\| \boldsymbol{D}_{G_k} \boldsymbol{\alpha}_{G_k} - \boldsymbol{r}_{G_k} \right\|_F^2 + \tau \left\| \boldsymbol{\alpha}_{G_k} \right\|_0 \quad (37)$$

where $\boldsymbol{D}_{G_k}$ is the self-adaptive learned dictionary from $\boldsymbol{r}_{G_k}$ using our proposed scheme described in subsection III.C. Obviously, Eq. (37) can also be considered as the sparse coding problem. Now we will show that the accurate solution of Eq. (37) can be achieved efficiently. With the definitions of $\{\boldsymbol{\alpha}_{G_k}\}$ and $\{\boldsymbol{\gamma}_{G_{r_k}}\}$, we get $\boldsymbol{x}_{G_k} = \boldsymbol{D}_{G_k} \boldsymbol{\alpha}_{G_k}$, $\boldsymbol{r}_{G_k} = \boldsymbol{D}_{G_k} \boldsymbol{\gamma}_{G_{r_k}}$. Due to the unitary property of $\boldsymbol{D}_{G_k}$, we have

$$\left\| \boldsymbol{D}_{G_k} \boldsymbol{\alpha}_{G_k} - \boldsymbol{D}_{G_k} \boldsymbol{\gamma}_{r_{G_k}} \right\|_F^2 = \left\| \boldsymbol{\alpha}_{G_k} - \boldsymbol{\gamma}_{r_{G_k}} \right\|_2^2. \quad (38)$$

With Eq. (38), the sub-problem (37) is equivalent to

$$\operatorname{argmin}_{\boldsymbol{\alpha}_{G_k}} \frac{1}{2} \left\| \boldsymbol{\alpha}_{G_k} - \boldsymbol{\gamma}_{r_{G_k}} \right\|_2^2 + \tau \left\| \boldsymbol{\alpha}_{G_k} \right\|_0. \quad (39)$$

Therefore, the closed-form solution of (39) is expressed as

$$\hat{\boldsymbol{\alpha}}_{G_k} = \operatorname{hard}(\boldsymbol{\gamma}_{r_{G_k}}, \sqrt{2\tau}) = \boldsymbol{\gamma}_{r_{G_k}} \odot 1(\operatorname{abs}(\boldsymbol{\gamma}_{r_{G_k}}) - \sqrt{2\tau}), \quad (40)$$

where $\operatorname{hard}(\cdot)$ denotes the operator of hard thresholding and $\odot$ stands for the element-wise product of two vectors. This process is applied for all $n$ groups to achieve $\hat{\boldsymbol{\alpha}}_G$, which is the final solution for $\boldsymbol{\alpha}_G$ sub-problem in Eq. (31).

*C. Summary of Proposed Algorithm*

So far, all issues in the process of handing the above two sub-problems have been solved. In fact, we acquire the efficient solution for each separated sub-problem, which enables the whole algorithm more efficient and effective. In light of all derivations above, a detailed description of the proposed algorithm for image restoration using GSR is provided in Table 1.

**Table 1:** A Complete Description of Proposed GSR Modeling for Image Restoration

**Input:** the observed image or measurement $y$ and the degraded operator $H$

**Initialization:** $t=0, b^{(0)}=0, \alpha_G^{(0)}=0, u^{(0)}, B_s, c, \lambda, \mu,$ ;

**Repeat**

    **if** $H$ is mask operator

        Update $u^{(t+1)}$ by Eq. (28);

    **else if** $H$ is blur operator

        Update $u^{(t+1)}$ by Eq. (28);

    **else if** $H$ is random projection operator

        Update $u^{(t+1)}$ by Eq. (30);

    **end if**

    $r^{(t+1)} = u^{(t+1)} - b^{(t)}; \tau = (\lambda K)/(\mu N);$

    **for** Each group $x_{G_k}$

        Construct dictionary $D_{G_k}$ by computing Eq. (16);

        Reconstruct $\hat{\alpha}_{G_k}$ by computing Eq. (40);

    **end for**

    Update $D_G^{(t+1)}$ by concatenating all $D_{G_k}$ ;

    Update $\hat{\alpha}_G^{(t+1)}$ by concatenating all $\hat{\alpha}_{G_k}$ ;

    Update $b^{(t+1)}$ by computing Eq. (26);

    $t \leftarrow t+1;$

**Until** maximum iteration number is reached

**Output:** Final restored image $\hat{x} = D_G \circ \hat{\alpha}_G$ .

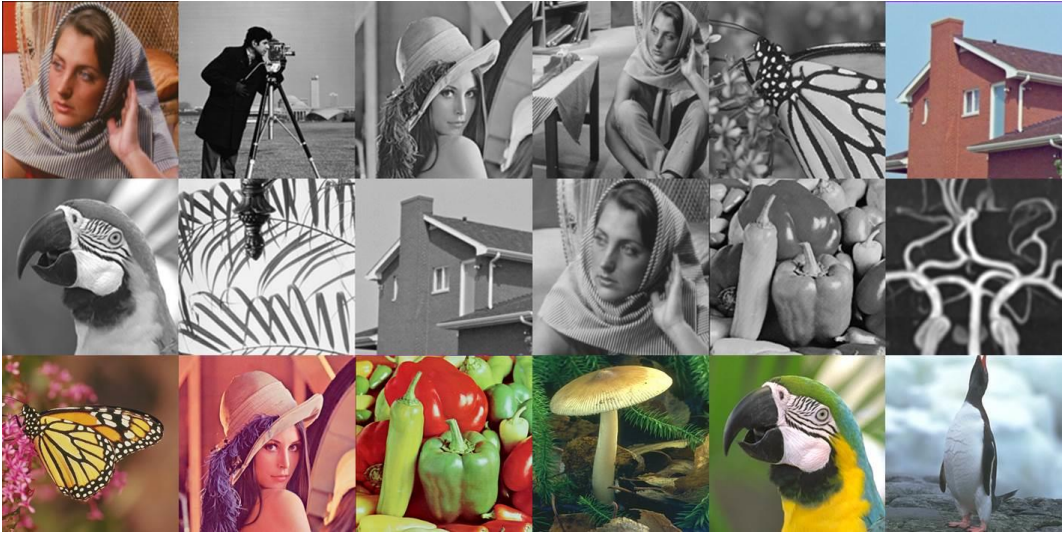

**Figure 4:** All experimental test images.

V. EXPERIMENTAL RESULTS

In this section, extensive experimental results are conducted to verify the performance of the proposed GSR for image restoration applications, which include image inpainting, image deblurring and image compressive sensing recovery. The parameter setting of GSR is as follows: the size of a group is set to be $64 \times 60$, with $\mathcal{B}_s$ being 64 and $c$ being 60. The width of overlapping between adjacent patches is 4 pixels, leading to the relationship $K = 240N$. The range of training window for constructing group, i.e., $L \times L$ is set to be $40 \times 40$. The parameters $\mu$ and $\lambda$ are set accordingly for different image restoration applications, which will be given below. All the experiments are performed in Matlab 7.12.0 on a Dell OPTIPLEX computer with Intel(R) Core(TM) 2 Duo CPU E8400 processor (3.00GHz), 3.25G memory, and Windows XP operating system.

To evaluate the quality of the reconstructed image, in addition to PSNR (Peak Signal to Noise Ratio, unit: dB), which is used to evaluate the objective image quality, a recently proposed powerful perceptual quality metric FSIM [45] is calculated to evaluate the visual quality. The higher FSIM value means the better visual quality. For color images, image restoration operations are only applied to the luminance component. All the experimental test images are given in Fig. 4. **Due to the limit of space, only parts of the experimental results are shown in this paper. Please enlarge and view the figures on the screen for better comparison.** Our Matlab code and all the experimental results can be downloaded at the website: http://idm.pku.edu.cn/staff/zhangjian/GSR/.

*A. Image Inpainting*

In this subsection, two interesting cases of image inpainting with different masks are considered, i.e., image restoration from partial random samples and text removal. For image inpainting application, $\mu = 0.0025$ and $\lambda = 0.082$.

The proposed GSR is compared with five recent representative methods for image inpainting: SKR (steering kernel regression) [6], NLTV [35], BPFA [48], HSR [10] and SAIST [5]. SKR [6] is a classic method that utilizes a steering kernel regression framework to characterize local structures for image restoration. NLTV [35] is an extension of traditional total variation (TV) with a nonlocal weight function. BPFA [48] employs a truncated beta-Bernoulli process to infer an appropriate dictionary for image recovery and exploits the spatial inter-relationships within imagery through the use of the Dirichlet and probit stick-breaking processes. HSR [10] combines the strength of local and nonlocal sparse representations under a systematic framework called Bayesian model averaging, characterizing local smoothness and nonlocal self-similarity simultaneously.

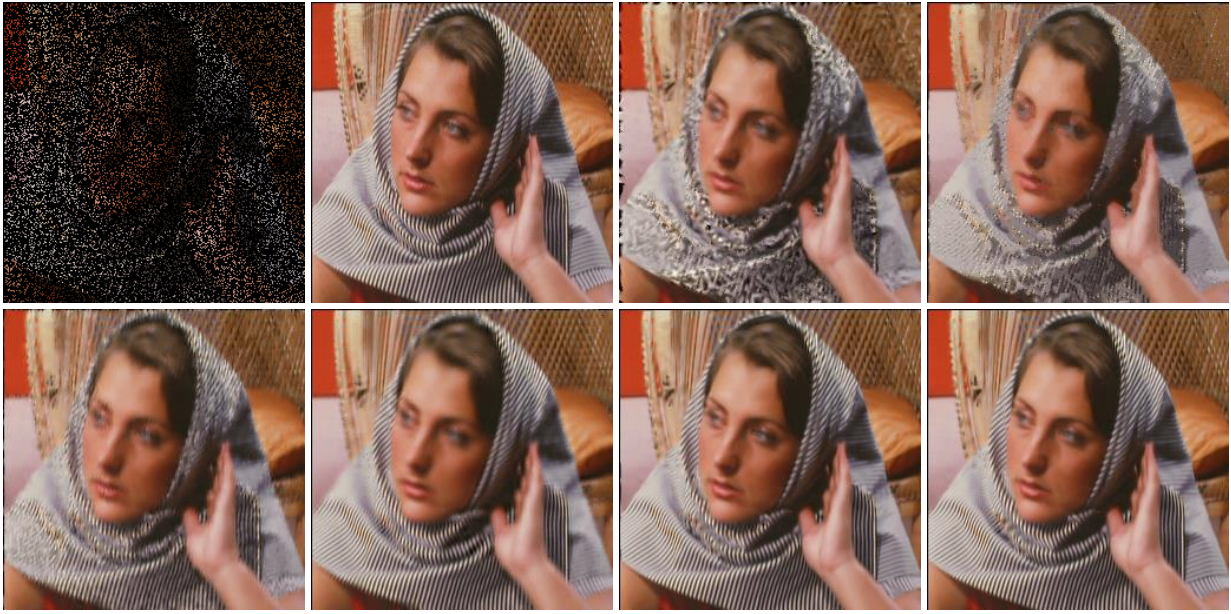

**Figure 5:** Visual quality comparison of image restoration from partial random samples for color image *Barbara*. From left to right and top to bottom: the degraded image with only 20% random samples available, original image, the recovered images by SKR [6] (PSNR=21.92dB; FSIM=0.8607), NLTV [35] (PSNR= 23.46dB; FSIM=0.8372), BPFA [48] (PSNR=25.70dB; FSIM=0.8926), HSR [10] (PSNR=28.83dB; FSIM=0.9273), SAIST [5] (PSNR=29.68dB; FSIM =0.9485) and the proposed GSR (PSNR=**31.32**dB; FSIM=**0.9598**).

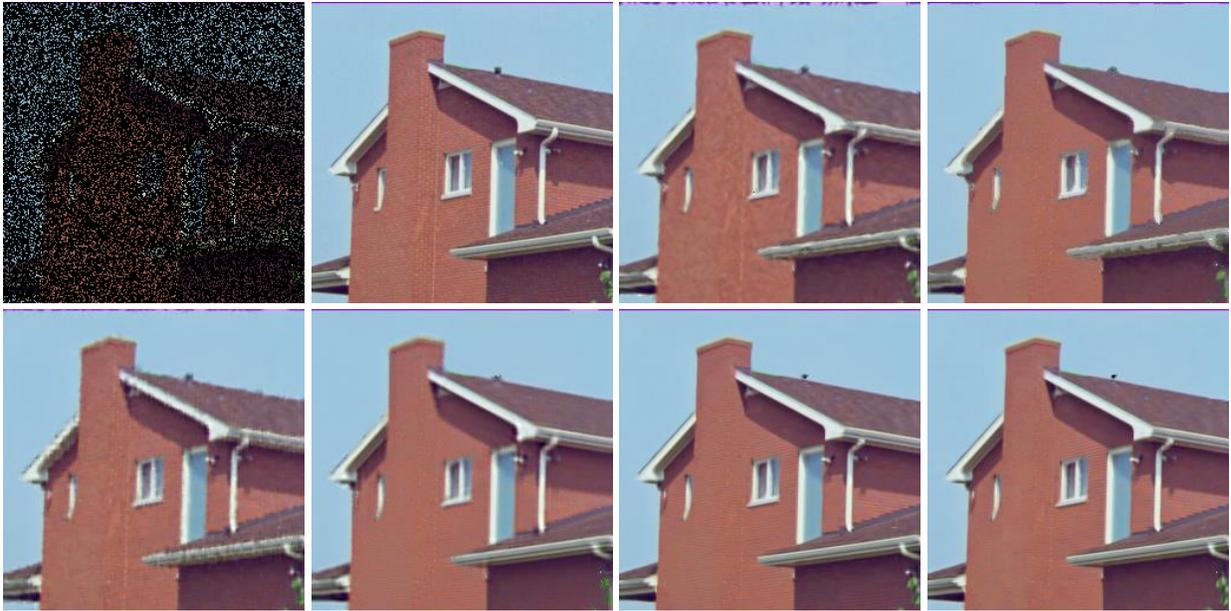

**Figure 6:** Visual quality comparison of image restoration from partial random samples for color image *House*. From left to right and top to bottom: the degraded image with only 20% random samples available, original image, the recovered images by SKR [6] (PSNR=30.40dB; FSIM=0.9198), NLTV [35] (PSNR=31.19dB; FSIM=0.9093), BPFA [48] (PSNR=30.89dB; FSIM=0.9111), HSR [10] (PSNR=32.35dB; FSIM=0.9255), SAIST [5] (PSNR=**35.73**dB; FSIM=**0.9615**) and the proposed GSR (PSNR=35.61dB; FSIM=0.9594).

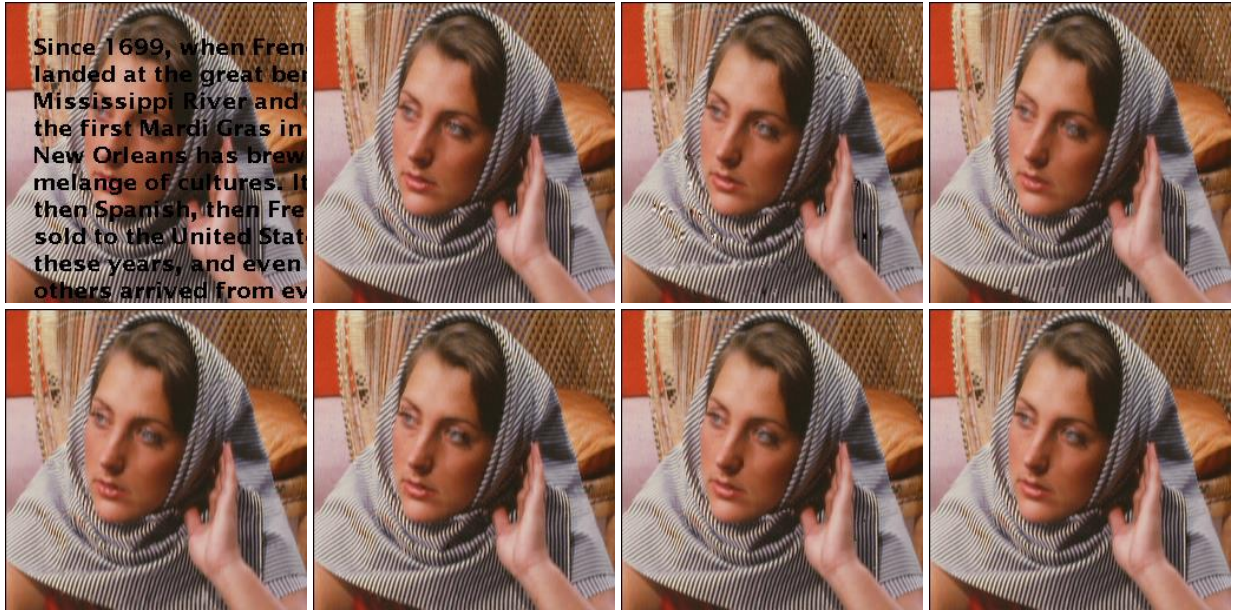

**Figure 7:** Visual quality comparison in the case of text removal for color image *Barbara*. From left to right and top to bottom: the masked image, original image, the recovered images by SKR [6] (PSNR=30.81dB; FSIM=0.9747), NLTV [35] (PSNR=32.60dB; FSIM=0.9749), BPFA [48] (PSNR=34.28dB; FSIM=0.9790), HSR [10] (PSNR=38.86dB; FSIM=0.9901), SAIST [5] (PSNR=39.00dB; FSIM=0.9915) and the proposed GSR (PSNR=**40.86**dB; FSIM=**0.9936**).

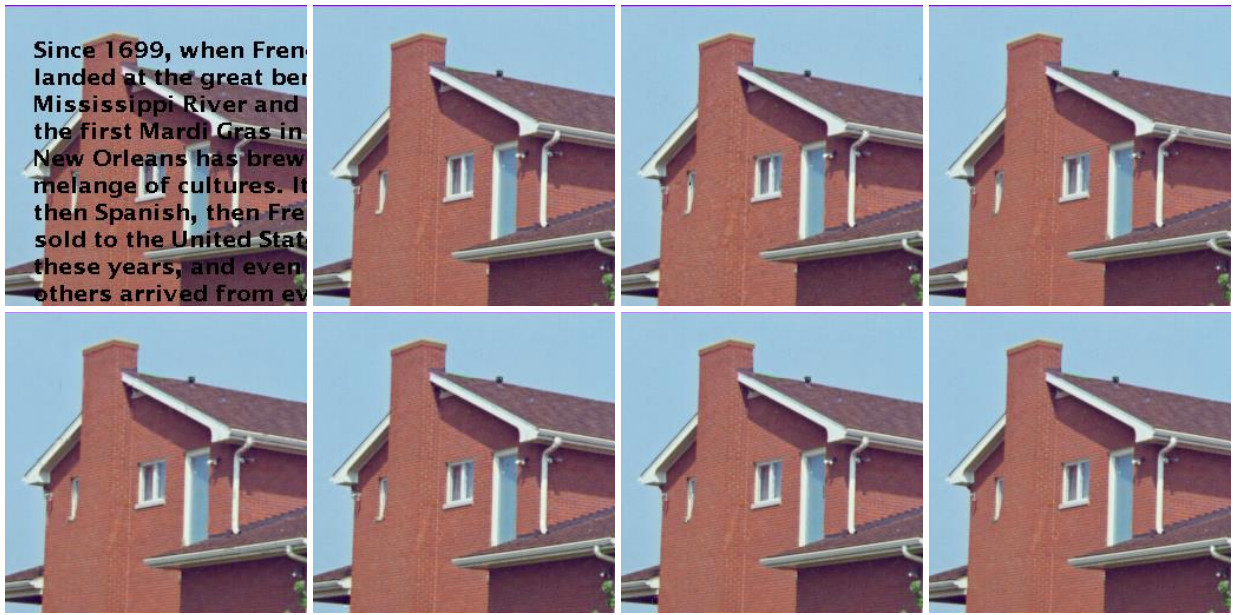

**Figure 8:** Visual quality comparison in the case of text removal for color image *House*. From left to right and top to bottom: the masked image, original image, the recovered images by SKR [6] (PSNR=38.65dB; FSIM=0.9850), NLTV [35] (PSNR=38.44dB; FSIM=0.9820), BPFA [48] (PSNR=39.01dB; FSIM=0.9818), HSR [10] (PSNR=42.06dB; FSIM=0.9913), SAIST [5] (PSNR=41.20dB; FSIM=0.9893) and the proposed GSR (PSNR=**42.51**dB; FSIM=**0.9916**).

The visual quality comparisons in the case of image restoration from only 20% random samples and in the case of text removal for two standard color test images are provided in Figs. 5~8. It is obvious to see that SKR and NLTV are good at capturing contour structures, but fail in recovering textures and produce blurred effects. BPFA is able to recover some textures, while generating some incorrect textures and some blurred effects due to less robustness with so small percentage of retaining samples for dictionary learning. HSR usually restores better textures than SKR, NLTV and BPFA. However, it often produces noticeable striped artifacts. We can observe that SAIST and the proposed GSR modeling can provide better restoration on both edges and textures than other competing methods. Concretely, for image *Barbara* which is full of textures, GSR achieves much better PSNR and FSIM than SAIST, with more image details and textures in both cases as shown in Fig. 5 and Fig. 7. For image *House* which is rich of edges, GSR achieves almost the same performance with SAIST in the case of image restoration from only 20% random sample (see Fig. 6), and achieves better result than SAIST in the case of text removal (see Fig. 8). Additional qualitative PSNR and FSIM results by our proposed GSR for image restoration from partial random samples on eight standard color images at different percentages of random samples are shown in Table 2.

**Table 2:** PSNR and FSIM results by GSR for image restoration from partial random samples at different percentages of random samples

| Data Percentage | Image | | | |
|---|---|---|---|---|
| | *Barbara* | *House* | *Parrots* | *Lena* |
| *20%* | 31.32/0.9598 | 35.61/0.9594 | 29.84/0.9530 | 34.12/0.9815 |
| *30%* | 34.42/0.9768 | 37.65/0.9745 | 33.31/0.9695 | 36.38/0.9893 |
| *50%* | 39.12/0.9906 | 41.61/0.9886 | 37.78/0.9853 | 39.83/0.9956 |
| Data Percentage | Image | | | |
| | *Butterfly* | *Mushroom* | *Penguin* | *Peppers* |
| *20%* | 30.31/0.9792 | 28.85/0.9020 | 33.23/0.9594 | 33.82/0.9811 |
| *30%* | 33.02/0.9888 | 30.87/0.9346 | 35.62/0.9732 | 35.47/0.9874 |
| *50%* | 37.26/0.9956 | 34.60/0.9702 | 39.28/0.9872 | 38.30/0.9941 |

*B. Image Deblurring*

In this subsection, two sets of experiments are conducted to verify the performance of the proposed GSR method for image deblurring. In the first set, two types of blur kernels, including a 9×9 uniform kernel and a Gaussian blur kernel, are exploited for simulation, with standard deviation of additive Gaussian noise $\sigma = \sqrt{2}$ (see Table 4). In the second set, six typical deblurring experiments (as shown in Table 3) with respect to four standard gray images, which have been presented in [14] [15] are provided.

The proposed GSR deblurring method is compared with four recently developed deblurring approaches, i.e., TVMM [9], L0_ABS [46], NCSR [15], and IDDBM3D [14]. TVMM [8] is a TV-based deblurring approach that can well reconstruct the piecewise smooth regions but often fails to recover fine image details. The L0_ABS [9] is a sparsity-based deblurring method exploiting a fixed sparse domain. IDDBM3D [10] method is an improved version of BM3D deblurring method [31]. NCSR proposed a centralized sparse constraint, which exploits the image nonlocal redundancy to reduce the sparse coding noise [15]. As far as we know, NCSR and IDDBM3D provide the current best image deblurring results in the literature.

**Table 3:** Six typical deblurring experiments with various blur PSFs and noise variances in the second set

| Scenario | PSF | $\sigma^2$ |
|---|---|---|
| 1 | $1/(1 + z_1^2 + z_2^2)$, $z_1, z_2 = -7, ..., 7$ | 2 |
| 2 | $1/(1 + z_1^2 + z_2^2)$, $z_1, z_2 = -7, ..., 7$ | 8 |
| 3 | 9×9 uniform | ≈ 0.3 |
| 4 | $[1\ 4\ 6\ 4\ 1]^T [1\ 4\ 6\ 4\ 1]/256$ | 49 |
| 5 | Gaussian with *std* = 1.6 | 4 |
| 6 | Gaussian with *std* = 0.4 | 64 |

The PSNR and FSIM results on six gray test images in the first set of experiments are reported in Table 4. For the case of 9×9 uniform kernel with noise $\sigma = \sqrt{2}$, $\mu = 0.0075$ and $\lambda = 0.554$, and for the case of Gaussian kernel with noise $\sigma = \sqrt{2}$, $\mu = 0.0125$ and $\lambda = 0.41$.

**Table 4:** PSNR and FSIM Comparisons for Image Deblurring in the First Set

| *Image* | *Barbara* | *Boats* | *House* | *C. Man* | *Peppers* | *Lena* | *Avg.* |
|---|---|---|---|---|---|---|---|
| **Uniform Kernel: 9×9, $\sigma = \sqrt{2}$** | | | | | | | |
| **TVMM** [9] | 26.00/0.8538 | 29.39/0.8978 | 32.47/0.9134 | 26.83/0.8674 | 28.77/0.8983 | 28.68/0.8937 | 28.69/0.8874 |
| **L0_ABS** [46] | 26.41/0.8692 | 29.77/0.9071 | 33.01/0.9241 | 27.12/0.8729 | 28.68/0.9078 | 28.76/0.9056 | 28.96/0.8978 |
| **IDDBM3D** [14] | 27.98/0.9014 | 31.20/0.9304 | 34.44/0.9369 | 28.56/0.9007 | 29.62/0.9200 | 29.70/0.9197 | 30.25/0.9182 |
| **NCSR** [15] | 28.10/0.9117 | 31.08/0.9294 | 34.31/**0.9415** | **28.62/0.9026** | **29.66**/0.9220 | 29.96/0.9254 | 30.29/0.9221 |
| **GSR** | **28.95/0.9227** | **31.34/0.9326** | **34.48**/0.9403 | 28.28/0.8937 | **29.66/0.9231** | **30.10/0.9281** | **30.47/0.9234** |
| **Gaussian Kernel: fspecial (Gaussian, 25, 1.6), $\sigma = \sqrt{2}$** | | | | | | | |
| **TVMM** [9] | 24.81/0.8435 | 30.44/0.9219 | 33.01/0.9139 | 27.04/0.8911 | 29.20/0.9237 | 30.72/0.9259 | 29.21/0.9034 |
| **L0_ABS** [46] | 24.78/0.8543 | 30.54/0.9297 | 33.07/0.9212 | 27.34/0.8955 | 28.88/0.9303 | 30.63/0.9361 | 29.21/0.9112 |
| **IDDBM3D** [14] | 27.19/0.8986 | 31.68/**0.9426** | 34.08/0.9359 | 28.17/**0.9136** | 29.99/**0.9373** | 31.45/0.9430 | 30.43/0.9285 |
| **NCSR** [15] | 27.91/0.9088 | 31.49/0.9371 | 33.63/0.9333 | **28.34**/0.9078 | 30.16/0.9331 | 31.26/0.9389 | 30.47/0.9265 |
| **GSR** | **28.26/0.9155** | **31.69**/0.9411 | **34.45/0.9420** | 27.78/0.9006 | **30.19**/0.9349 | **31.47/0.9463** | **30.64/0.9301** |

From Table 4, we can see that the proposed GSR achieves highly competitive performance compared with other leading deblurring methods. L0_ABS produces slightly higher average PSNR and FSIM than TVMM, while GSR outperforms L0_ABS by 1.5 dB and 1.4 dB for the uniform blur and Gaussian blur, respectively. One can observe that IDDBM3D, NCSR and GSR produce very similar results, and obtain significant PSNR/FSIM improvements over other competing methods. In average, GSR outper-forms IDDBM3D and NCSR by (0.22 dB, 0.21 dB) and (0.18 dB, 0.17 dB) for the two blur kernels, respectively. The visual comparisons of the deblurring methods are shown in Figs. 9~10, from which one can observe that the GSR model produces cleaner and sharper image edges and textures than other competing methods.

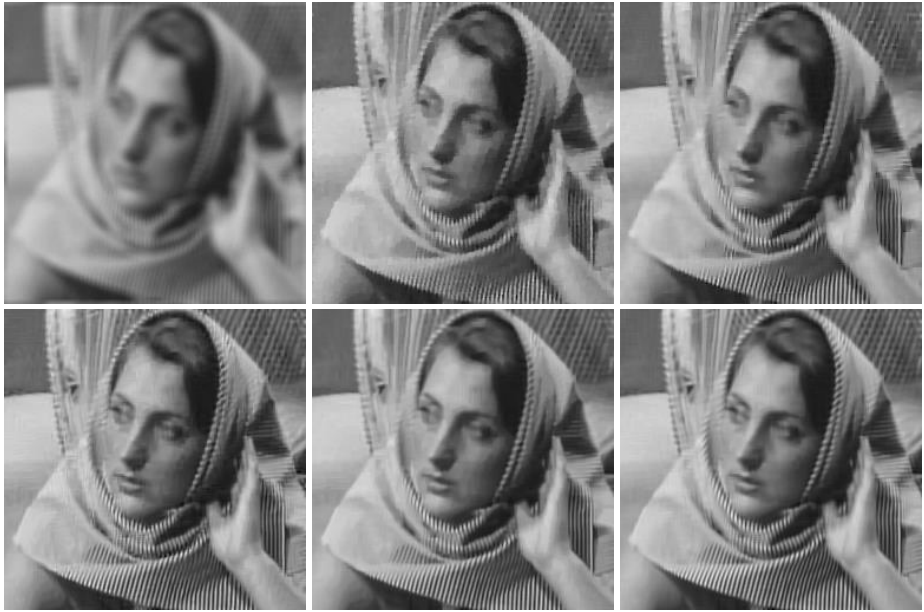

**Figure 9:** Visual quality comparison of image deblurring on gray image *Barbara* (256×256). From left to right and top to bottom: noisy and blurred image (uniform kernel: 9×9, $\sigma=\sqrt{2}$), the deblurred images by TVMM [9] (PSNR=26.00dB; FSIM=0.8538), L0_ABS [46] (PSNR=26.41dB; FSIM=0.8692), NCSR [15] (PSNR=28.10dB; FSIM=0.9117), IDDBM3D [14] (PSNR=27.98dB; FSIM= 0.9014) and the proposed GSR (PSNR=**28.95**dB; FSIM=**0.9227**).

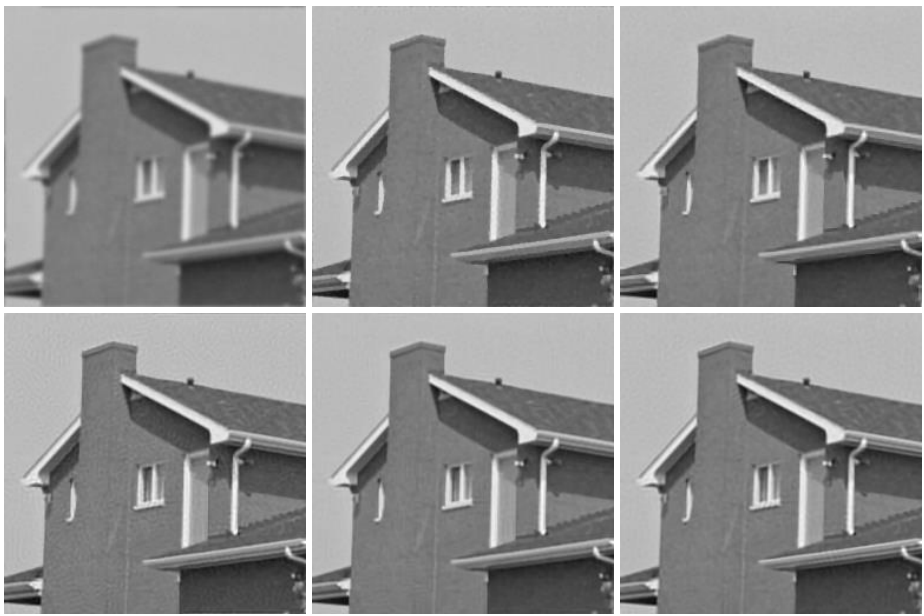

**Figure 10:** Visual quality comparison of image deblurring on gray image *House* (256×256). From left to right and top to bottom: noisy and blurred image (Gaussian kernel: $\sigma = \sqrt{2}$), the deblurred images by TVMM [9] (PSNR=33.01dB; FSIM=0.9139), L0_ABS [46] (PSNR=33.07dB; FSIM=0.9212), NCSR [15] (PSNR=33.63dB; FSIM=0.9333), IDDBM3D [14] (PSNR=34.08dB; FSIM= 0.9359) and the proposed GSR (PSNR=**34.45**dB; FSIM=**0.9420**).

Table 5: Comparison of the ISNR (dB) Deblurring Results in the Second Set

|  | Scenario | | | | | | Scenario | | | | | |
| --- | --- | --- | --- | --- | --- | --- | --- | --- | --- | --- | --- | --- |
|  | 1 | 2 | 3 | 4 | 5 | 6 | 1 | 2 | 3 | 4 | 5 | 6 |
| **Method** | *Cameraman* (256×256) | | | | | | *House* (256×256) | | | | | |
| **BSNR** | 31.87 | 25.85 | 40.00 | 18.53 | 29.19 | 17.76 | 29.16 | 23.14 | 40.00 | 15.99 | 26.61 | 15.15 |
| **Input PSNR** | 22.23 | 22.16 | 20.76 | 24.62 | 23.36 | 29.82 | 25.61 | 25.46 | 24.11 | 28.06 | 27.81 | 29.98 |
| **TVMM [9]** | 7.41 | 5.17 | 8.54 | 2.57 | 3.36 | 1.30 | 7.98 | 6.57 | 10.36 | 4.12 | 4.54 | 2.44 |
| **L0_ABS [46]** | 7.70 | 5.55 | 9.10 | 2.93 | 3.49 | 1.77 | 8.40 | 7.12 | 11.06 | 4.55 | 4.80 | 2.15 |
| **IDDBM3D [14]** | **8.85** | **7.12** | **10.45** | **3.98** | 4.31 | **4.89** | 9.95 | 8.55 | 12.89 | 5.79 | 5.74 | 7.13 |
| **NCSR [15]** | 8.78 | 6.69 | 10.33 | 3.78 | **4.60** | 4.50 | 9.96 | 8.48 | 13.12 | 5.81 | 5.67 | 6.94 |
| **GSR** | 8.39 | 6.39 | 10.08 | 3.33 | 3.94 | 4.76 | **10.02** | **8.56** | **13.44** | **6.00** | **5.95** | **7.18** |
| **Method** | *Lena* (512×512) | | | | | | *Barbara* (512×512) | | | | | |
| **BSNR** | 29.89 | 23.87 | 40.00 | 16.47 | 27.18 | 15.52 | 30.81 | 24.79 | 40.00 | 17.35 | 28.07 | 16.59 |
| **Input PSNR** | 27.25 | 27.04 | 25.84 | 28.81 | 29.16 | 30.03 | 23.34 | 23.25 | 22.49 | 24.22 | 23.77 | 29.78 |
| **TVMM [9]** | 6.36 | 4.98 | 7.47 | 3.52 | 3.61 | 2.79 | 3.10 | 1.33 | 3.49 | 0.41 | 0.75 | 0.59 |
| **L0_ABS [46]** | 6.66 | 5.71 | 7.79 | 4.09 | 4.22 | 1.93 | 3.51 | 1.53 | 3.98 | 0.73 | 0.81 | 1.17 |
| **IDDBM3D [14]** | 7.97 | 6.61 | 8.91 | 4.97 | 4.85 | 6.34 | 7.64 | 3.96 | 6.05 | 1.88 | 1.16 | 5.45 |
| **NCSR [15]** | 8.03 | 6.54 | 9.25 | 4.93 | 4.86 | 6.19 | 7.76 | 3.64 | 5.92 | 2.06 | 1.43 | 5.50 |
| **GSR** | **8.24** | **6.76** | **9.43** | **5.17** | **4.96** | **6.57** | **8.98** | **4.80** | **7.15** | **2.19** | **1.58** | **6.20** |

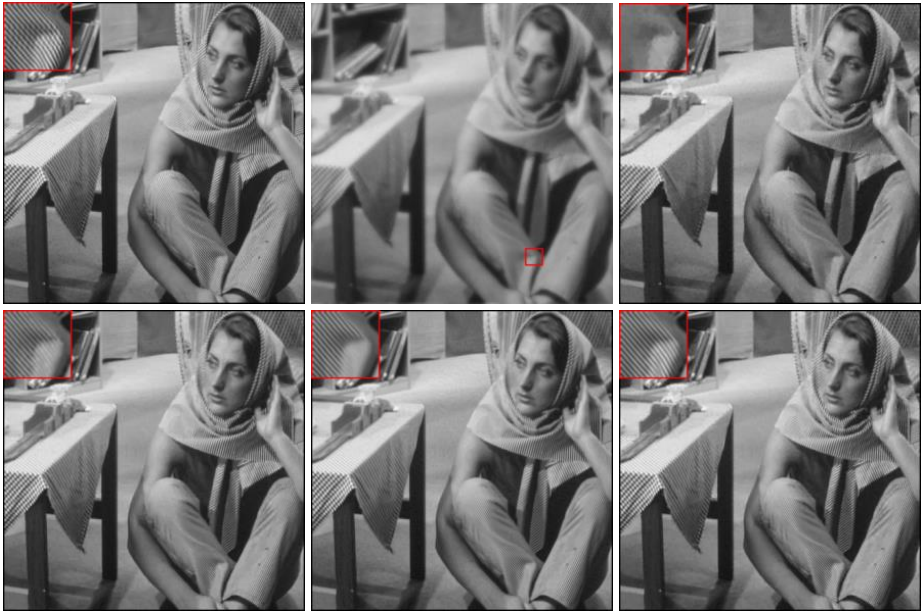

**Figure 11:** Visual quality comparison of image deblurring on image *Barbara* (512×512). From left to right and top to bottom: original image, noisy and blurred image (scenario 2), the deblurred images by TVMM [8] (PSNR=24.58dB; FSIM=0.9576), IDDBM3D [10] (PSNR=27.21dB; FSIM=0.9699), NCSR [11] (PSNR=26.89dB; FSIM=0.9669) and the proposed GSR (PSNR=**28.05**dB; FSIM=**0.9738**).

Table 5 lists the comparison of ISNR results for six typical deblurring experiments in the second set. It is clear to observe that GSR achieves the highest ISNR results in the most cases, as labeled in bold. In particular, for image *Barbara* (512×512) with rich textures, GSR outperforms current state-of-the-art methods NCSR and IDDBM3D more than 1 dB in the scenarios 1, 2, 3, with more textures and clearer edges than other competing methods, as shown in Fig. 11. More visual results can be found in the website of this paper.

*C. Image Compressive Sensing Recovery*

From many fewer acquired measurements than suggested by the Nyquist sampling theory, CS theory demonstrates that a signal can be reconstructed with high probability when it exhibits sparsity in some domain, which has greatly changed the way engineers think of data acquisition. More specifically, suppose that we have an image $\boldsymbol{x} \in \mathbb{R}^N$ and its measurement $\boldsymbol{y} \in \mathbb{R}^M$, namely, $\boldsymbol{y} = \boldsymbol{Hx}$. Here, $\boldsymbol{H}$ is an $M \times N$ measurement matrix such that $M$ is much smaller than $N$. The purpose of image CS recovery is to recover $\boldsymbol{x}$ from $\boldsymbol{y}$ with measurement rate, denoted by ratio, equal to $M/N$. For image CS recovery application, $\mu = 0.0025$ and $\lambda = 0.082$.

In our simulations, the CS measurements are obtained by applying a Gaussian random projection matrix to the original image signal at block level, i.e., block-based CS with block size of $32 \times 32$ [47]. GSR is compared with four representative CS recovery methods in literature, i.e., wavelet method (DWT), total variation (TV) method [41], multi-hypothesis (MH) method [40], collaborative sparsity (CoS) method [42], which deal with image signals in the wavelet domain, the gradient domain, the random projection residual domain, and the hybrid space-transform domain, respectively. It is worth emphasizing that MH and CoS are known as the current state-of-the-art algorithms for image CS recovery.

**Table 6:** PSNR and FSIM Comparisons with Various CS Recovery Methods (dB)

| Ratio | Algorithms | House | Barbara | Leaves | Monarch | Parrot | Vessels | Avg. |
|---|---|---|---|---|---|---|---|---|
| 20% | DWT | 30.70/0.9029 | 23.96/0.8547 | 22.05/0.7840 | 24.69/0.8155 | 25.64/0.9161 | 21.14/0.8230 | 24.70/0.8494 |
| | TV [41] | 31.44/0.9051 | 23.79/0.8199 | 22.66/0.8553 | 26.96/0.8870 | 26.6/0.9018 | 22.04/0.8356 | 25.59/0.8675 |
| | MH [40] | 33.60/0.9370 | 31.09/0.9419 | 24.54/0.8474 | 27.03/0.8707 | 28.06/0.9332 | 24.95/0.8756 | 28.21/0.9010 |
| | CoS [42] | 34.34/0.9326 | 26.60/0.8742 | 27.38/0.9304 | 28.65/0.9171 | 28.44/0.9282 | 26.71/0.9214 | 28.69/0.9259 |
| | **GSR** | **36.78/0.9618** | **34.59/0.9703** | **29.90/0.9499** | **29.55/0.9236** | **30.60/0.9512** | **31.58/0.9599** | **32.17/0.9528** |
| 30% | DWT | 33.60/0.9391 | 26.26/0.8980 | 24.47/0.8314 | 27.23/0.8628 | 28.03/0.9445 | 24.82/0.8924 | 27.40/0.8947 |
| | TV [41] | 33.75/0.9384 | 25.03/0.8689 | 25.85/0.9092 | 30.01/0.9279 | 28.71/0.9329 | 25.13/0.9011 | 28.08/0.9131 |
| | MH [40] | 35.54/0.9546 | 33.47/0.9614 | 27.65/0.8993 | 29.18/0.9003 | 31.20/0.9529 | 29.36/0.9360 | 31.07/0.9341 |
| | CoS [42] | 36.69/0.9592 | 29.49/0.9267 | 31.02/0.9606 | 31.38/0.9449 | 30.39/0.9490 | 31.35/0.9664 | 31.72/0.9511 |
| | **GSR** | **38.93/0.9761** | **36.92/0.9811** | **33.82/0.9731** | **33.17/0.9558** | **33.95/0.9671** | **36.33/0.9841** | **35.53/0.9729** |
| 40% | DWT | 35.69/0.9576 | 28.53/0.9327 | 26.82/0.8741 | 29.58/0.9011 | 30.06/0.9588 | 29.53/0.9467 | 30.03/0.9285 |
| | TV [41] | 35.56/0.9574 | 26.56/0.9088 | 28.79/0.9442 | 32.92/0.9538 | 30.54/0.9530 | 28.14/0.9441 | 30.42/0.9436 |
| | MH [40] | 37.04/0.9676 | 35.20/0.9727 | 29.93/0.9276 | 31.07/0.9217 | 33.21/0.9651 | 33.49/0.9677 | 33.32/0.9537 |
| | CoS [42] | 38.46/0.9724 | 32.76/0.9618 | 33.87/0.9744 | 33.98/0.9637 | 32.55/0.9627 | 33.95/0.9784 | 34.26/0.9689 |
| | **GSR** | **40.60/0.9836** | **38.99/0.9877** | **37.02/0.9846** | **36.07/0.9714** | **36.44/0.9780** | **40.24/0.9922** | **38.23/0.9829** |

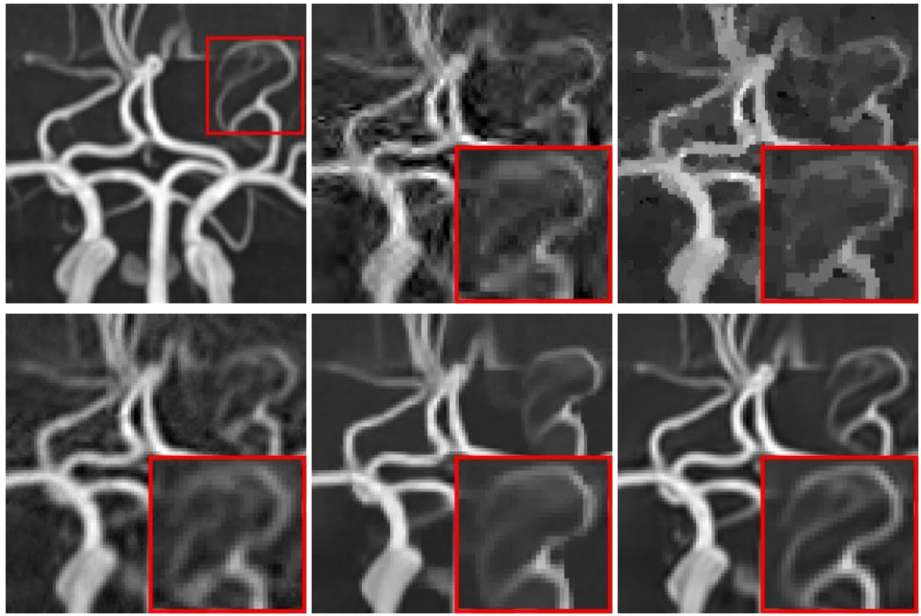

**Figure 12:** Visual quality comparison of image CS recovery on gray image *Vessels* in the case of ratio = 20%. From left to right: original image, the CS recovered images by DWT (PSNR=21.14dB; FSIM=0.8230), TV [41] (PSNR=22.04dB; FSIM=0.8356), MH [40] (PSNR=24.95dB; FSIM=0.8756), CoS [42] (PSNR= 26.71dB; FSIM=0.9214) and the proposed GSR (PSNR= **31.58**dB; FSIM=**0.9599**).

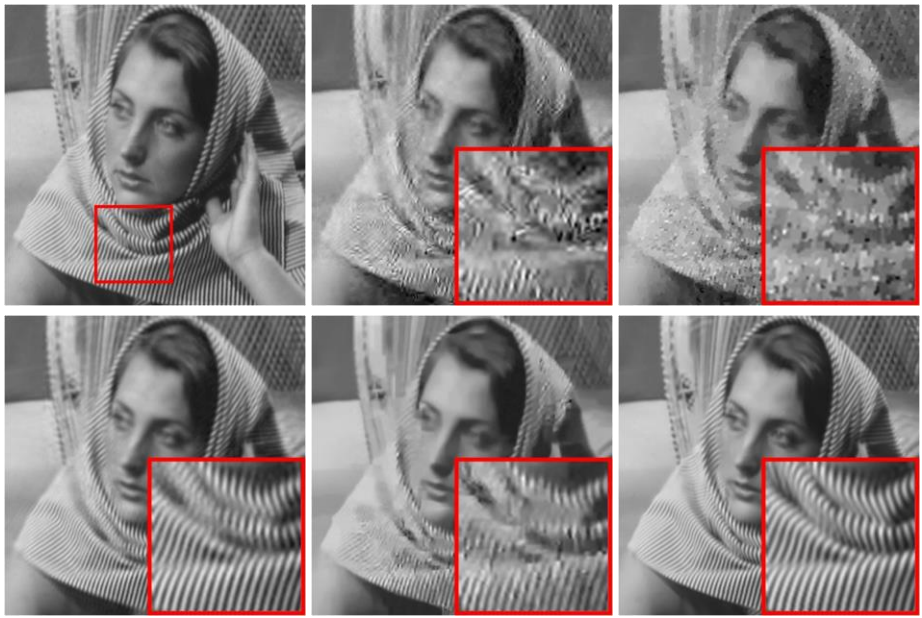

**Figure 13:** Visual quality comparison of image CS recovery on gray image *Barbara* in the case of ratio = 20%. From left to right and top to bottom: original image, the CS recovered images by DWT (PSNR=23.96dB; FSIM=0.8547), TV [41] (PSNR =23.79dB; FSIM =0.8199), MH [40] (PSNR=31.09dB; FSIM=0.9419), CoS [42] (PSNR= 26.60dB; FSIM=0.8742) and the proposed GSR (PSNR=**34.59**dB; FSIM =**0.9703**).

The PSNR and FSIM comparisons for six gray test images in the cases of 20%, 30%, and 40% measurements are provided in Table 6. GSR achieves the highest PSNR and FSIM among the six comparative algorithms over all the cases, which can improve roughly 7.9 dB, 7.3 dB, 4.4 dB, and 3.7 dB on average, in comparison with DWT, TV, MH, CoS, respectively, greatly improving existing CS recovery results.

Some visual results of the recovered images by various algorithms are presented in Figs. 12~13. Obviously, DWT and TV generate the worst perceptual results. The CS recovered images by MH and CoS possess much better visual quality than those of DWT and TV, but still suffer from some undesirable artifacts, such as ringing effects and lost details. The proposed algorithm GSR not only eliminates the ringing effects, but also preserves sharper edges and finer details, showing much clearer and better visual results than the other competing methods. Our work also offers a fresh and successful instance to corroborate the CS theory applied for natural images.

### D. Effect of Number of Best Matched Patches

This subsection will give the detailed description about how sensitive the performance is affected by $c$, which is the number of best matched patches.

To investigate the sensitivity of $c$, experiments with respect to various $c$, ranging from 20 to 120, in the case of image inpainting and image deblurring for three test images are conducted. The performance comparison with various $c$ is shown in Fig. 14. From Fig.14, it is concluded that the performance of our proposed algorithm is not quite sensitive to $c$ because all the curves are almost flat. The highest performance for each case is usually achieved with $c$ in the range [40, 80]. Therefore, in this paper, $c$ is empirically set to be 60.

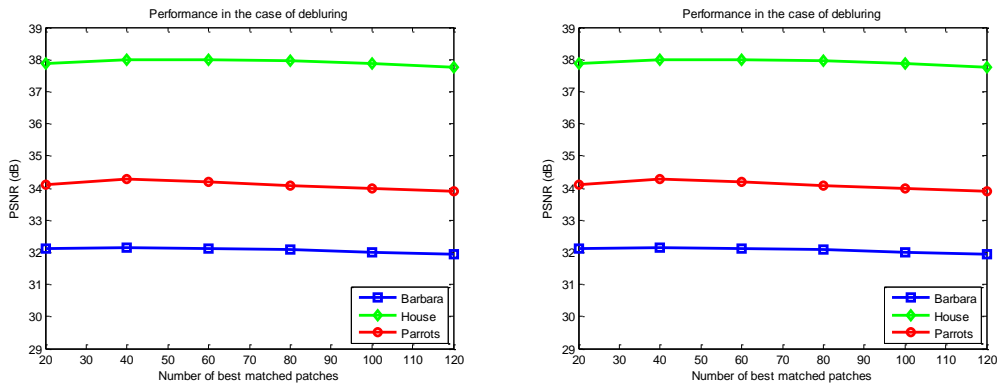

**Figure 14:** Performance comparison with various $c$ for three test images. From left to right: in the case of image text removal and in the case of image deblurring with 9×9 uniform blur kernel and $\sigma = 0.5$.

### E. Effect of Sparsity Parameter

This subsection gives some discussion about how sensitive the performance is affected by the sparsity parameter $\lambda$.

To investigate the effect of the sparsity parameter $\lambda$ for the performance, two scenarios of deblurring experiments are conducted with various blur kernels and noise variances, i.e., scenario 4 and scenario 5 in Table 3. Fig. 15 provides ISNR (dB) evolution with respect to $\lambda$ in the cases of scenario 4 (PSF = $[1\ 4\ 6\ 4\ 1]^T [1\ 4\ 6\ 4\ 1]$, $\sigma = 7$) and scenario 5 (PSF = Gaussian with std = 1.6, $\sigma = 2$) for two test images. From Fig. 15, three conclusions can be observed. First, as expected, there is an optimal $\lambda$ that achieves the highest ISNR by balancing image noise suppression with image details preservation (see Fig. 16(c)). That means, if $\lambda$ is set too small, the image noise can't be suppressed (see Fig. 16(b)); if $\lambda$ is set too large, the image details will be lost (see Fig. 16(d)). Second, in each case, the optimal $\lambda$ for each test image is almost the same. For instance, in the case of $\sigma = 7$, the optimal $\lambda$ is 12.2, and in the case of $\sigma = 2$, the optimal $\lambda$ is 0.8. This is very important for parameter optimization, since the optimal $\lambda$ in each case can be determined by only one test image and then applied to other test images. Third, it is obvious to see that $\lambda$ has a great relationship with $\sigma$, i.e., a larger $\sigma$ corresponds to a larger $\lambda$.

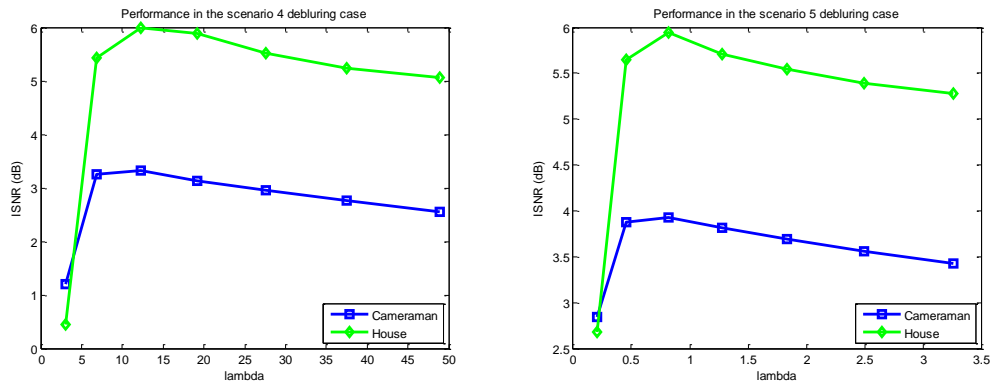

**Figure 15:** PSNR evolution with respect to sparsity parameter $\lambda$ in the cases of in the cases of scenario 4 (PSF = $[1\ 4\ 6\ 4\ 1]^T [1\ 4\ 6\ 4\ 1]$, $\sigma = 7$) and scenario 5 (PSF = Gaussian with std = 1.6, $\sigma = 2$) for two test images.

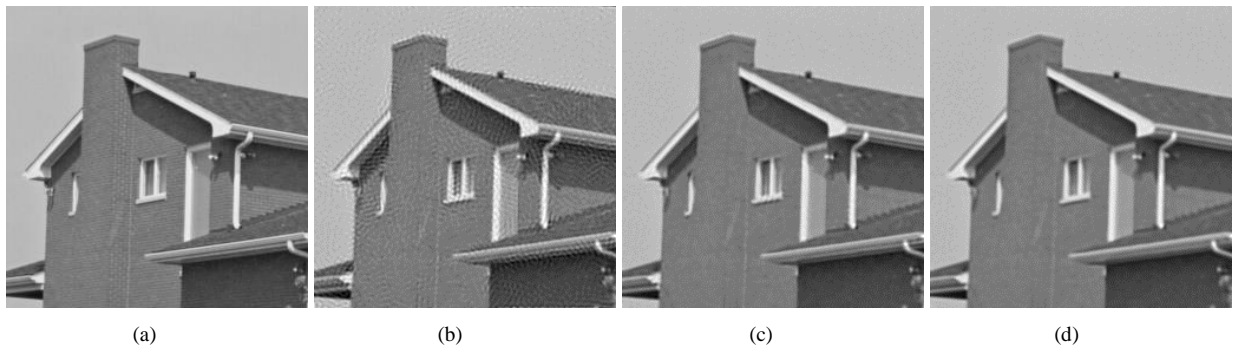

**Figure 16:** Visual quality comparison of proposed algorithm with various $\lambda$ in the case of scenario 5 (PSF = Gaussian with std = 1.6, $\sigma = 2$) with respect to image *House*. (a) Original image; (b) Deblurred result with $\lambda = 0.2$, ISNR=2.68dB; (c) Deblurred result with $\lambda = 0.8$, ISNR=**5.95**dB; (d) Deblurred result with $\lambda = 3.2$, ISNR=5.28dB.

## F. Algorithm Complexity and Computational Time

The complexity of GSR is provided as follows. Assume that the number of image pixels is *N*, that the average time to compute similar patches for each reference patch is $T_s$. The SVD of each group $x_{G_k}$ with size of $B_s \times c$ is $\mathcal{O}(B_s \times c^2)$. Hence, the total complexity of GSR for image restoration is $\mathcal{O}(N(B_s c^2 + T_s))$. For a 256×256 image, the proposed algorithm GSR requires about 8~9 minutes for image inpainting, 6~7 minutes for image deblurring and 7~8 minutes for CS recovery, on an Intel Core2 Duo 2.96G PC under Matlab R2011a environment.

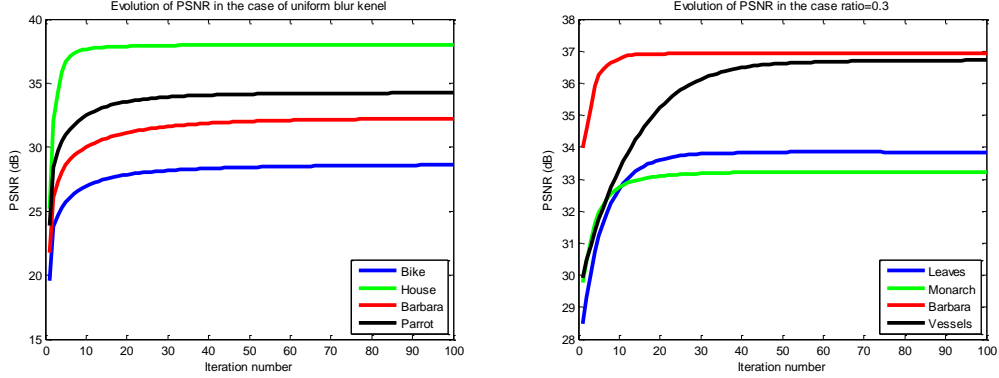

**Figure 17:** Stability of the proposed algorithm. From left to right: progression of the PSNR (dB) results achieved by proposed GSR for test images with respect to the iteration number in the cases of image deblurring with uniform blur kernel and image CS recovery with ratio=0.3.

## G. Algorithm Stability

Since the objective function (12) is non-convex, it is difficult to give its theoretical proof for global convergence. Here, we only provide empirical evidence to illustrate the stability of the proposed GSR. Take the cases of image CS recovery and image deblurring as examples. Fig. 17 plots the evolutions of PSNR versus iteration numbers for test images in the cases of image deblurring with uniform blur kernel and CS recovery with ratio=0.3. It is observed that with the growth of iteration number, all the PSNR curves increase monotonically and ultimately become flat and stable, exhibiting good stability of the proposed GSR model.

## H. Comparison between $\ell_0$ and $\ell_1$ Minimization

In order to make a comparison between $\ell_0$ and $\ell_1$ minimization, split Bregman iteration (SBI) is also used to solve Eq. (21). The only difference from solving Eq. (12) described in Table 1 is $\hat{\alpha}_{G_k}$ in Eq. (40) is computed by the operator of soft thresholding, rather than hard thresholding. Take the cases of image deblurring with uniform blur kernel for two images *Barbara* and *Parrot* as examples. Fig. 18 plots their progression curves of the PSNR (dB) results achieved by proposed GSR-driven $\ell_0$ and $\ell_1$ minimization with respect to the iteration number. The result achieved by GSR-driven $\ell_0$ minimization with SBI is denoted by SBI+L0 (blue solid line), while the result achieved by GSR-driven $\ell_1$ minimization with SBI is denoted by SBI+L1 (green dotted line). It is obvious that SBI+L0 has better performance than SBI+L1 with more than 1.5 dB on average, which fully demonstrates and the superiority of $\ell_0$ minimization (Eq. (12)) over $\ell_1$ minimization (Eq. (21)), and validates the effectiveness of our proposed approach to solve Eq. (12) again. Our study also assures the feasibility of using the $\ell_0$ minimization for image restoration problems.

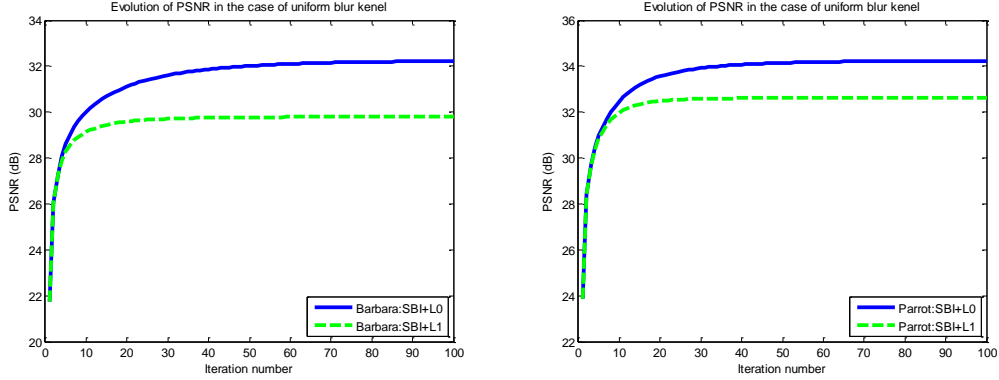

**Figure 18:** Comparison between GSR-driven $\ell_0$ and $\ell_1$ minimization solved by SBI. From left to right: progression of the PSNR (dB) results achieved by proposed GSR-driven $\ell_0$ and $\ell_1$ minimization with respect to the iteration number for images *Barbara* and *Parrot* in the cases of image deblurring with uniform blur kernel.

*I. Comparison between SBI and IST*

In our previous work [47], the convex optimization approach iterative shrinkage/thresholding (IST) is utilized to solve our proposed GSR-driven $\ell_0$ minimization for image CS recovery. Here, we make a comparison between SBI and IST. Take the cases of image CS recovery with ratio=0.3 for two images *Monarch* and *Leaves* as examples. Fig. 19 plots their progression curves of the PSNR (dB) results achieved by solving GSR-driven $\ell_0$ minimization with SBI and IST. The result achieved by $\ell_0$ minimization with SBI is denoted by SBI+L0 (red solid line), while the result achieved by $\ell_0$ minimization with IST is denoted by IST+L0 (black dotted line). Obviously, SBI is more efficient and effective to solve our proposed GSR-driven $\ell_0$ minimization problem than IST.

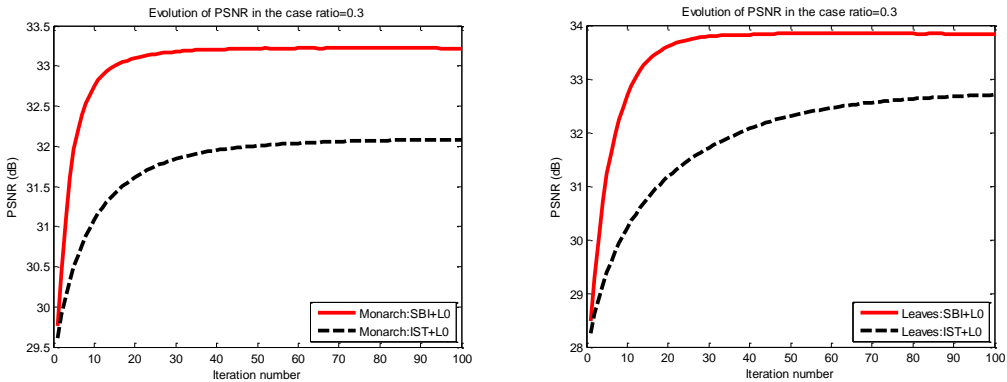

**Figure 19:** Comparison between SBI and IST for solving GSR-driven $\ell_0$ minimization. From left to right: progression of the PSNR (dB) results achieved by proposed GSR-driven $\ell_0$ minimization with respect to the iteration number for images *Monarch* and *Leaves* in the cases of image CS recovery with ratio=0.3.

## VI. Conclusions

This paper establishes a novel and general framework for high-quality image restoration using group-based sparse representation (GSR) modeling, which sparsely represents natural images in the domain of group, and explicitly and effectively characterizes the intrinsic local sparsity and nonlocal self-similarity of natural images simultaneously in a unified manner. An effectual self-adaptive group dictionary learning technique with low complexity is designed. To achieve high sparsity degree and high recovery quality, this paper proposes to exploit the convex optimization algorithms to solve the non-convex $\ell_0$ minimization problem directly. Our study not only assures the feasibility of using the $\ell_0$ minimization for image restoration problems, but also demonstrates the superiority of the $\ell_0$ minimization over the $\ell_1$ minimization, which is very interesting and surprising. Experimental results on three applications: image inpainting, deblurring and CS recovery have shown that the proposed GSR achieves significant performance improvements over many current state-of-the-art schemes and exhibits good stability. It is worth emphasizing that GSR greatly improves existing CS recovery results, which will promote further research and development of CS theory applied in natural images. Future work includes the extensions of GSR on a variety of applications, such as image deblurring with mixed Gaussian and impulse noise, and video restoration and so on.

## APPENDIX A

Proof of **Theorem 1**:

Due to the assumption that each $e(j)$ is independent, we obtain that each $e(j)^2$ is also independent. Since $E[e(j)]=0$ and $\mathrm{Var}[e(j)]=\sigma^2$, we have the mean of each $e(j)^2$, that is

$$E[e(j)^2] = \mathrm{Var}[e(j)] + [E[e(j)]]^2 = \sigma^2, \quad j=1,...,N. \tag{41}$$

By invoking *Law of Large Numbers* in probability theory, for any $\varepsilon>0$, it yields $\lim_{N\to\infty} P\{|\frac{1}{N}\sum_{j=1}^{N} e(j)^2 - \sigma^2| < \frac{\varepsilon}{2}\} = 1$, i.e.,

$$\lim_{N\to\infty} P\{|\tfrac{1}{N}\|\boldsymbol{x}-\boldsymbol{r}\|_2^2 - \sigma^2| < \tfrac{\varepsilon}{2}\} = 1, \tag{42}$$

Further, let $\boldsymbol{x}_G, \boldsymbol{r}_G$ denote the concatenation of all $\boldsymbol{x}_{G_k}$ and $\boldsymbol{r}_{G_k}$, $k=1,2,...,n$, respectively, and denote each element of $\boldsymbol{x}_G - \boldsymbol{r}_G$ by $\boldsymbol{e}_G(i)$, $i=1,...,K$. Due to the assumption, we conclude that $\boldsymbol{e}_G(i)$ is independent with zero mean and variance $\sigma^2$.

Therefore, the same manipulations with Eq. (40) applied to $\boldsymbol{e}_G(i)^2$ lead to $\lim_{K\to\infty} P\{|\frac{1}{K}\sum_{i=1}^{K} \boldsymbol{e}_G(i)^2 - \sigma^2| < \frac{\varepsilon}{2}\} = 1$, namely,

$$\lim_{K\to\infty} P\{|\tfrac{1}{K}\sum_{k=1}^{n}\|\boldsymbol{x}_{G_k}-\boldsymbol{r}_{G_k}\|_F^2 - \sigma^2| < \tfrac{\varepsilon}{2}\} = 1. \tag{43}$$

Considering Eqs. (42) and (43) together, we prove Eq. (34). □